\documentclass[journal]{IEEEtran}
\usepackage{amsmath,amsthm}
\usepackage{amssymb}
\usepackage{amsfonts}
\usepackage{graphicx}
\usepackage{float} 
\usepackage{subfigure}
\usepackage{tikz,tikz-3dplot}
\usepackage{tikzscale}
\usepackage{xcolor}
\usepackage{bm}
\usepackage{dsfont}
\usepackage{stfloats}
\usepackage[flushleft]{threeparttable}
\usepackage[numbers,sort&compress]{natbib}

\usepackage[switch]{lineno}
\usepackage{mathrsfs}
\usepackage{bbm}

\usepackage{svg}
\usepackage[colorlinks]{hyperref}
\hypersetup{colorlinks,breaklinks,linkcolor=blue,urlcolor=blue,anchorcolor=blue,citecolor=blue}
\usepackage{booktabs}
\usepackage{tabularx}
\usepackage{multirow}
\usepackage{lscape}

\usepackage{epstopdf}
\usepackage{booktabs}
\usepackage{caption}
\usepackage{enumitem}
\usepackage[flushleft]{threeparttable}

\usepackage{float}
\usepackage{microtype}
\usepackage{placeins}
\usepackage{xcolor}
\definecolor{deepblue}{rgb}{0,0,255}

\usepackage[defaultcolor=blue]{changes}

\usepackage{algorithm}
\usepackage{algorithmic}

\usepackage{threeparttable}
\usepackage{dcolumn}
\newcolumntype{d}[1]{D{.}{.}{#1}}

\ifCLASSINFOpdf
\else
\fi

\usepackage{xcolor}
\definecolor{darkblue}{rgb}{0.0,0.5,0.5}

\definecolor{myblue}{rgb}{0,0,255}

\definecolor{myred}{rgb}{255,0,0}

\begin{document}

\title{Bayesian Deep Learning Approach for Real-time Lane-based Arrival Curve Reconstruction at Intersection using License Plate Recognition Data}

\author{\IEEEauthorblockN{Yang He, Chengchuan An, Jiawei Lu, Yao-Jan Wu, Zhenbo Lu, and Jingxin Xia}

\thanks{This work was supported in part by the National Natural Science Foundation of China under Grant 52272309 and 52202398, and in part by the International Science and Technology Cooperation Project of Jiangsu Province under Grant BZ2023015 (Corresponding authors: Jingxin Xia)}.
\thanks{Yang He, Chengchuan An, Zhenbo Lu, and Jingxin Xia are with the Intelligent Transportation System Research Center, Southeast University, Nanjing, 211189, China (e-mail: yanghe@seu.edu.cn, ccan@seu.edu.cn, luzhenbo@seu.edu.cn, xiajingxin@seu.edu.cn).}
\thanks{Jiawei Lu is with the H. Milton Stewart School of Industrial and Systems Engineering,  Georgia Institute of Technology, Atlanta, Georgia, USA (email: lujiaweiwk@gmail.com).}
\thanks{Yao-Jan Wu is with the Department of Civil and Architectural Engineering and Mechanics, the University of Arizona, Tucson, AZ 85721, USA (yaojan@arizona.edu).}

}


\maketitle

\begin{abstract}
  The acquisition of real-time and accurate traffic arrival information is of vital importance for proactive traffic control systems, especially in partially connected vehicle environments. License plate recognition (LPR) data that record both vehicle departures and identities are proven to be desirable in reconstructing lane-based arrival curves in previous works.
  Existing LPR data-based methods are predominantly designed for reconstructing historical arrival curves. For real-time reconstruction of multi-lane urban roads, it is pivotal to determine the lane choice of real-time link-based arrivals, which has not been exploited in previous studies.
  In this study, we propose a Bayesian deep learning approach for real-time lane-based arrival curve reconstruction, in which the lane choice patterns and uncertainties of link-based arrivals are both characterized. Specifically, the learning process is designed to effectively capture the relationship between partially observed link-based arrivals and lane-based arrivals, which can be physically interpreted as lane choice proportion. Moreover, the lane choice uncertainties are characterized using Bayesian parameter inference techniques, minimizing arrival curve reconstruction uncertainties, especially in low LPR data matching rate conditions.
  Real-world experiment results conducted in multiple matching rate scenarios demonstrate the superiority and necessity of lane choice modeling in reconstructing arrival curves.
  
\end{abstract}


\IEEEdisplaynontitleabstractindextext

\IEEEpeerreviewmaketitle

\section{Introduction}

Arrival curves depict time-stamped cumulative vehicle arrivals across signal cycles, offering a more realistic and informative traffic arrival profile compared to the mean or cyclic arrival rates \cite{ni2015traffic, an2021lane, li2023traffic, an2024one}.
Real-time arrival curves are valuable in deriving timely performance measures (e.g., real-time queue length \cite{zheng2017estimating, an2018real, lee2019advanced, hao2024stochastic}) and hence optimizing signal timing \citep{li2018adaptive, yao2019dynamic} and connected and automated vehicle (CAV) trajectory \citep{xu2020trajectory, yao2020integrated, amini2021optimizing}, etc.
However, estimating lane-specific real-time arrival curves on multi-lane urban roads is still a challenging task due to the complicated lane choice behaviors associated with different movements.

Previous studies extensively exploited data from fixed-location detectors and mobile sensors to acquire traffic arrival information. Fixed advance detectors, installed at a fixed distance upstream of the stop bar, are commonly used to detect arrivals. 
However, when the vehicular queue exceeds the detector location, known as the queue-over-detector (QOD) condition, newly arriving vehicles can not be detected, leading to under-estimated arrivals \cite{dobrota2022novel}. Despite further studies imposing uniform assumptions on arrivals during QOD, the estimated arrivals are approximated.
Mobile sensors, such as floating cars and connected vehicles, offer arrival data with broader spatial coverage. Despite the expected increase, the penetration rate remains low for a relatively long term. As a result, only a few probe vehicle samples are available at intersections within a signal cycle, limiting their application primarily to characterizing the cyclic arrival patterns \cite{tan2021cumulative}.

In recent years, License Plate Recognition (LPR) technology has seen widespread adoption in cities across China and globally, paving the way for innovative applications, such as queue length estimation \cite{luo2019queue,tang2022lane, shao2018license, hao2024stochastic}, dynamic vehicular demand estimation \citep{rao2018origin,mo2020estimating}, and commuting pattern analysis \citep{yao2021understanding}, etc. License Plate Recognition (LPR) data integrates the strengths of both fixed-location and mobile sensors while offering superior features.
At the intersection level, LPR functions similarly to fixed detectors by identifying lane-based vehicle departures, but with the additional capability of recording each vehicle's unique license plate. At the network level, akin to mobile sensors, LPR enables vehicle tracking across multiple intersections by matching their license plate, yet it surpasses mobile sensors with substantially higher penetration rates. With the tracking ability and high penetration, LPR data have facilitated great advancements in effectively estimating cyclic arrival rates \cite{an2021lane, li2023traffic, an2024one} and reconstructing the more challenging arrival curves \cite{zhan2015lane, mo2017speed}.

The acquisition of arrival curves is equivalent to determining the cumulative arrival indices of all vehicles. Ideally, for two adjacent intersections monitored by LPR sensors, all these indices can be determined by matching license plates from the upstream to the downstream intersection. 
However, in practice, the arrival curve reconstruction problem is raised as partial unmatched vehicles exist owing to recognition failures, lack of upstream LPR sensors, mid-block traffic merging, etc. To tackle this, the existing approaches mainly formulated an interpolation-based method to infer the arrival indices of unmatched vehicles based on matched vehicles. For example, \citet{zhan2015lane} customized a cyclic arrival function-based Gaussian Process (GP) model to interpolate the arrival indices of unmatched vehicles while characterizing arrival uncertainties.
\citet{mo2017speed} further improved the GP model by mending the erroneous license plates, enhancing reconstruction robustness in undesirable matching conditions.

However, applying the interpolation methods to real-time reconstruction presents challenges.
These methods rely on previously matched vehicles to reconstruct arrival curves \cite{zhan2015lane, mo2017speed}, but the lane choices of real-time link arrivals are often undetermined. 
As a result, the intended direction of the vehicles traveling on the road remains unknown, leading to inaccurate real-time reconstruction.
Therefore, modeling the lane choices of real-time arriving vehicles is pivotal for real-time reconstruction. Moreover, lane choice behaviors are often stochastic owing to different driving preferences. For example, through-going drivers may choose either the lane with the shortest queues or the one requiring the fewest manipulations. Consequently, characterizing the uncertainties in lane choice becomes essential, especially in data-limited scenarios where these behavioral patterns are difficult to observe.

This study proposes a Bayesian deep learning approach for real-time lane-based arrival curve reconstruction using LPR data, in which lane choice patterns and uncertainties are both characterized. Specifically, the learning process is designed to capture the relationships between partially observed link-based arrivals and lane-specific arrivals, which can be physically interpreted as lane choice proportion.  Moreover, the lane choice uncertainties are modeled to minimize arrival curve reconstruction uncertainties, especially in low LPR data matching rate conditions.

The contributions of this study are summarized as follows

\begin{enumerate}

    \item A Bayesian deep learning approach is proposed to reconstruct real-time lane-based arrival curves using LPR data, in which lane choice patterns are characterized. Specifically, the learning process is designed to capture the relationship between partially observed link-based arrivals and lane-specific arrivals.
    
    \item Building on lane choice learning, Bayesian parameter inference techniques are used to robustly characterize lane choice uncertainties. This development effectively minimizes the arrival curve reconstruction uncertainties in low LPR data matching rate (e.g., 10 $\%$) conditions.
    
    \item The proposed approach is validated on a real-world dataset using both deterministic and probabilistic evaluation metrics. Experimental results indicate the superiority of the proposed lane-choice learning-based method and the necessity of modeling lane-choice uncertainties. 

\end{enumerate}

\section{Literature review}

According to the sensing technologies acquiring traffic arrivals, the existing approaches can be broadly divided into two categories, fixed-location detector-based and mobile sensor-based.
Fixed detector-based approaches capture the arrival data mainly by detecting the vehicle's presence using advanced detectors installed at a fixed distance upstream of the stop bar at the intersection \cite{day2010evaluation}.
However, when the queue tail exceeds the detector, known as the queue-over-detector (QOD) condition, new arrivals can not be detected due to continuous occupation of the detector \citep{an2018real,lee2019advanced}. The detector occupation leads to underestimated traffic arrivals, especially in over-saturated conditions when queues can not dissipate in one signal cycle.
To address this problem, researchers \cite{zheng2014use, dobrota2022novel} have assumed a uniform distribution of traffic arrival during the QOD condition, though this is only an approximation of actual arrivals.

With the popularization of ride-hailing services and the emergence of connected vehicle technologies, massive vehicle trajectory data from mobile sensors have gained great attention in traffic state monitoring and performance evaluation \citep{zhan2016citywide, zhao2019various}. 
Existing studies, assuming stationary arrival pattern during a historical period, have utilized historical data from multiple signal cycles to estimate cyclic arrival rates through various methods such as constant global index difference \citep{hao2013vehicle}, EM-solved maximum likelihood estimation \citep{zheng2017estimating, zhang2019cycle}, and kernel density estimation \citep{tan2021cumulative}. 
Prior studies have indicated that sufficient probe samples, typically at least one or two per signal cycle, are required to capture more refined arrival profiles \citep{li2017real, wen2023inferring}.
However, despite expected increases, the penetration rate of mobile sensors remains low in the current and near future. The unbalanced spatial distribution of mobile sensors further exacerbates these challenges, limiting their application to cyclic arrival pattern analysis.

As a special type of fixed-location detector, LPR systems not only record vehicle passing timestamps and lanes but also register informative vehicle license plate identities. 
In addition, LPR integrates the strength of mobile sensors
by enabling vehicle tracking across the network through license plate matching. Further, it presents substantially higher penetration rates and spatial coverage than mobile sensors, making it desirable for both estimating cyclic arrival rates \citep{an2021lane, an2024one} and reconstructing more challenging arrival curves \citep{zhan2015lane, mo2017speed}.
The existing arrival curve reconstruction methods mainly extracted lane-based arrival indices by matching the vehicle's license plate and then formulated interpolation-based models to infer the indices of those unmatched vehicles. 
\citet{zhan2015lane} customized an arrival function-based Gaussian process (GP) method to interpolate the cumulative arrival indices of unmatched vehicles while capturing the arrival uncertainties. The piece-wise linear arrival function is used by assuming a cyclic arrival pattern which may not always guaranteed in practice.
\citet{mo2017speed} further improved the GP model by developing a data-mending approach to correct the erroneously recognized license plate, providing more matched vehicles and enhancing model robustness in limited data conditions.
In summary, the existing studies have effectively leveraged LPR data to reconstruct arrival curves, accounting for their stochastic natures and practical constraints.  
Despite the great success, these interpolation-based approaches still rely on historical matched vehicles regarding real-time reconstruction. The informative real-time link-based arrival data available from LPR systems are not exploited, which weakens their capability in real-time applications.

\begin{figure}[t]
  \centering
  \subfigure[Problem statement]{ 
  \includegraphics[width=3.5in]{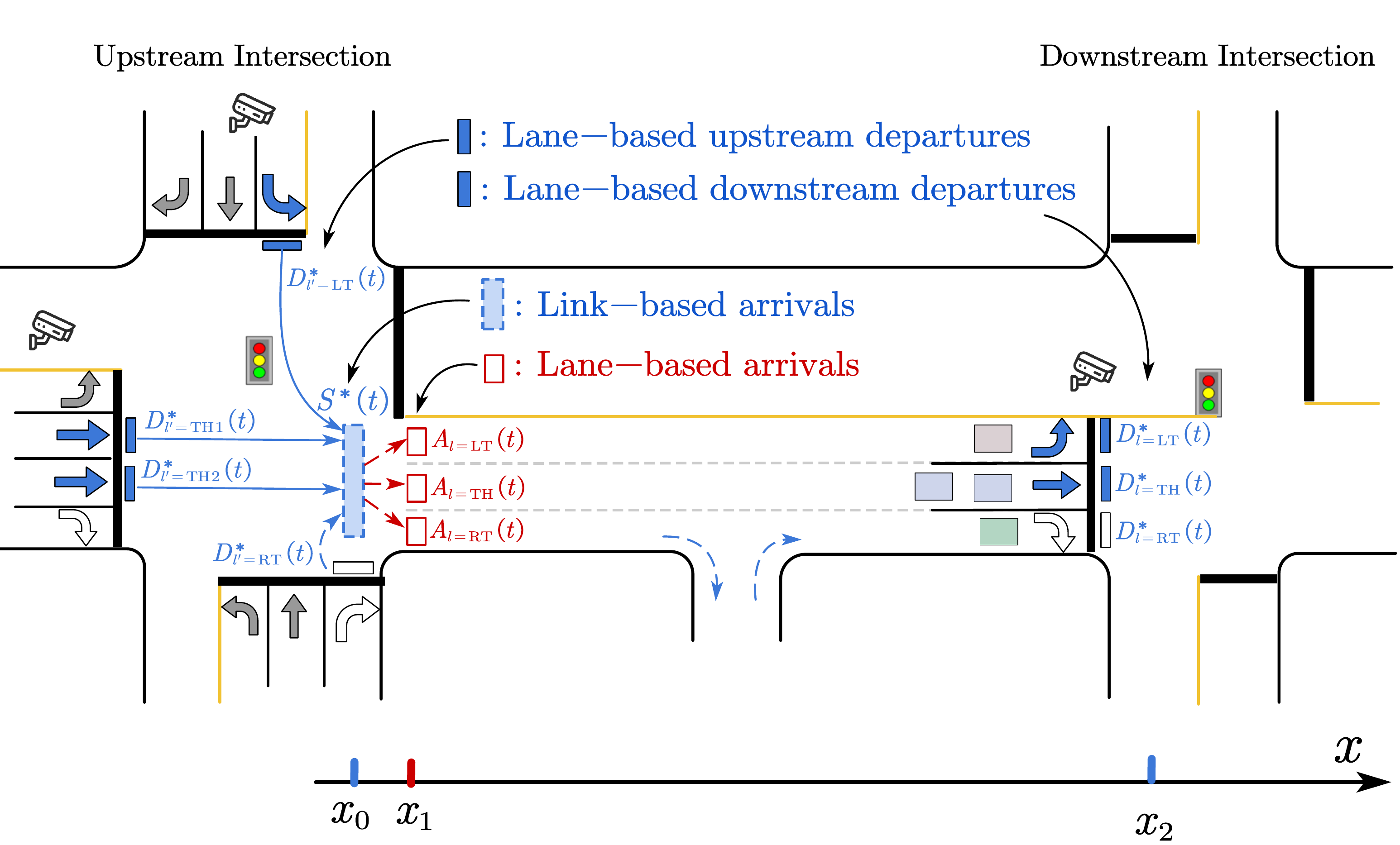}}
  \subfigure[Cumulative arrival and departure curves]{ 
  \includegraphics[width=3.5in]{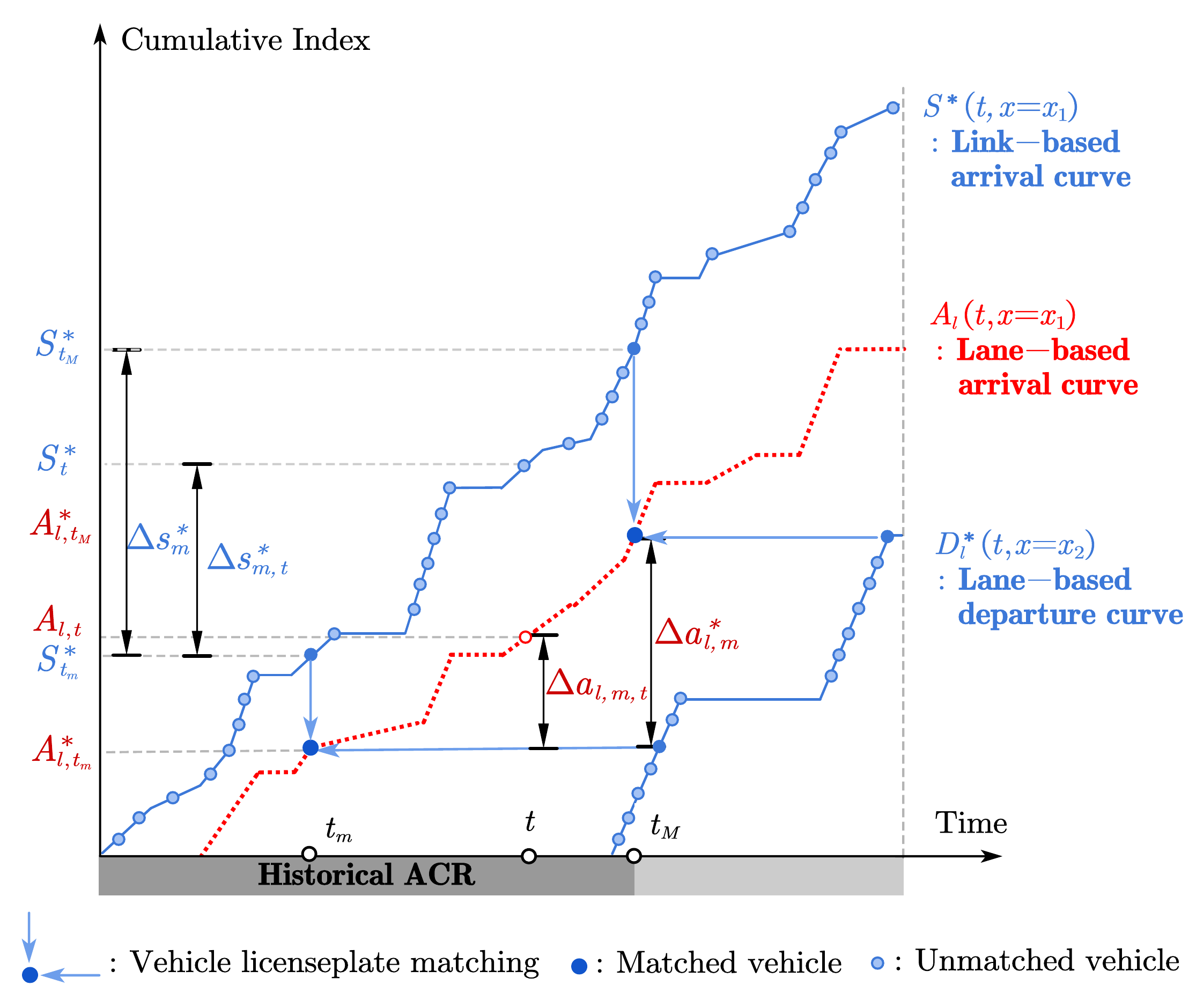}}
  \caption{Problem statement and the relationship between cumulative departure and arrival curves. }
  \label{fig: problem statement}
\end{figure}

\section{Problem Statement}
\label{problem statement}

\subsection{Lane-based Arrival Curve Reconstruction (ACR)} \label{subsec: Lane-based ACR task}

Fig. \ref{fig: problem statement}(a) describes a common scenario of the license plate recognition (LPR) data collection at two adjacent intersections. LPR cameras are usually installed to monitor the vehicle departures of left-turn and through traffic under the protective phase (i.e., hollow rectangles in Fig. \ref{fig: problem statement}(a)), while the right-turn traffic under the permitted phase (i.e., hollow rectangles in Fig. \ref{fig: problem statement}(a)) is not monitored. 
By indexing the recognized departing vehicles recorded in LPR data, the lane-based cumulative departure curves at upstream and downstream intersections can be obtained.  
By matching the license plates of departed vehicles from the upstream to the downstream intersection, the lane-based arrival indices of partial vehicles can be determined, as shown in Fig. \ref{fig: problem statement}(b).
The arrival curve reconstruction problem is raised as not all the vehicles can be matched owing to recognition failures, lack of upstream LPR sensors, mid-block traffic merging, etc. 
There are several assumptions used in this study

\begin{itemize}
    \item \textbf{Assumption 1}: The vehicle travel time within the intersection is ignored. As a result, the traffic departures from the stop-line of the upstream approaching lane (i.e., $x=x_0$) are approximated to the link-based traffic arrivals at the upstream section of the target link (i.e., $x=x_1$),
    
    \item \textbf{Assumption 2}: The mid-block merging traffic arrivals are regarded as the link-based traffic arrivals at the upstream section of the link (i.e., $x=x_1$) since they will eventually depart from the downstream lanes.
    
    \item \textbf{Assumption 3}: The upstream arriving traffic arrivals that diverge to side streets in the mid-block section are not considered in this study, as they do not pass the downstream lanes.
\end{itemize}

To obtain the lane-based traffic arrival curve, there are four types of vehicle cumulative curves of concern:
\begin{enumerate}
        \item Lane-based upstream departure curves at the section of upstream approaching lane $x_0$, ${\color[RGB]{60, 120, 216} D_{l'}^{*}\left( t,x=x_0 \right) }$, where $l'$ is the upstream lane heading to the target link, e.g., $l'\in \left\{ \mathrm{LT},\mathrm{TH}1,\mathrm{TH}2,\mathrm{RT} \right\}$ in Fig. \ref{fig: problem statement}.
        \item Link-based arrival curve at the upstream section of target link $x_1$, ${\color[RGB]{60, 120, 216} S^{*}\left( t,x=x_1 \right) }$,
	\item Lane-based  departure curves at downstream section $x_3$, ${\color[RGB]{60, 120, 216} D_{l}^{*}\left( t,x=x_3 \right) }$, where $l$ is the downstream lane, e.g., $l\in \left\{ \mathrm{LT},\mathrm{TH},\mathrm{RT} \right\}$ in Fig. \ref{fig: problem statement}.
	\item Lane-based arrival curve at the upstream section $x_1$, ${\color[RGB]{204, 0, 0} A_{l}^{}\left( t,x=x_1 \right) }$, where $l$ is the downstream lane, e.g., $l \in \left\{ \mathrm{LT},\mathrm{TH},\mathrm{RT} \right\}$ in Fig. \ref{fig: problem statement}.
\end{enumerate}

Based on Assumption 1, the upstream link-based arrival curve is approximated to the sum of upstream lane-based departures that head to the target link, i.e., 
\begin{align}
S^*\left( t, x=x_1\right) =\sum_{l'}{D_{l'}^{*}\left( t,  x=x_0\right)}, 
\label{eq: link-based arrivals}
\end{align}
where $l'$ is the upstream lane heading to the target link, e.g., $l'\in \left\{ \mathrm{LT},\mathrm{TH}1,\mathrm{TH}2,\mathrm{RT} \right\}$ in Fig. \ref{fig: problem statement}.
This study aims to estimate real-time lane-based traffic arrival curves at the upstream section of the target link  ${\color[RGB]{204, 0, 0} A_{l}^{}\left( t,x=x_1 \right) }$ based on the upstream link-based arrival curve ${\color[RGB]{60, 120, 216} S^{*}\left( t,x=x_1 \right) }$ and downstream lane-based departure curves ${\color[RGB]{60, 120, 216} D_{l}^{*}\left( t,x=x_3 \right) }$ using LPR data from upstream and downstream intersections. 
Furthermore, based on the estimated upstream arrivals, the traffic arrivals at arbitrary sections between the upstream and downstream can be determined using a free-flow speed, i.e.,
\begin{align}
	A_l\left( t,x \right) =A_l\left( t-\frac{x-x_1}{v_f},x_1 \right) ,x_1\leqslant x\leqslant x_2,
\label{eq: arrival curve at arbitrary section}
\end{align}
where $v_f$ is the free-flow speed. For example, upstream arrival curves combined with stochastic free-flow speeds are employed to obtain the pseudo arrival curves at the downstream section (i.e., $x=x_2$) as shown in Fig. \ref{fig: problem statement}(b), which is used to calculate the queue profile at intersection \cite{hao2024stochastic}.

For the sake of simplicity, we hide the spatial index of the arrival and departure curves, since the symbol contains spatial information. For example, the lane-based arrival curve at the upstream sections $A_{l}^{}\left( t,x=x_1 \right) $ is simply denoted as $A_{l}^{}\left( t \right)$.

\begin{figure*}[t]
  \begin{center}
  \includegraphics[width=7in]{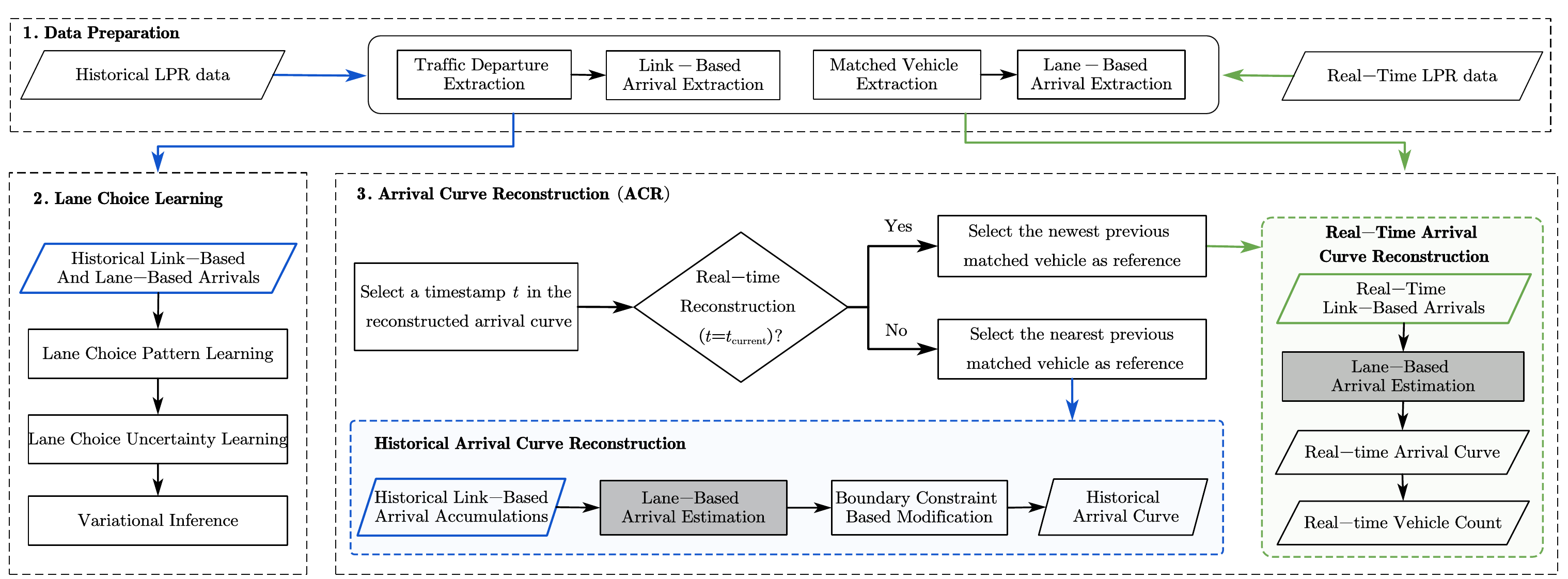}\\
  \caption{Overview of the proposed lane choice learning-based arrival curve reconstruction framework.}\label{fig: framework}
  \end{center}
\end{figure*}

\subsection{Cumulative Departure Curves and Matched Vehicles} \label{subsec: cumulative departure curves and matched vehicles}

According to the LPR records, we can obtain the accurate and complete timestamp sequences for all vehicles that depart from the stop bar on each monitored lane of both upstream and downstream intersections. By ordering the vehicle departing timestamp $t_i$, the corresponding cumulative departure index  $D_{l, t_i}^{*}$ of each vehicle can be determined. 
The cumulative departure curve $D_{l'}^{*}\left( t\right) $ can be obtained by connecting departure observations  $\left( t_i, D_{l, t_i}^{*} \right) , i=1,2,...,D$.

The matched vehicles could be obtained through vehicle license plate matching between upstream and downstream lanes using vehicle departure records, as shown in Fig. \ref{fig: problem statement}(b). 
Specifically, a vehicle is regarded as a matched vehicle $m$ if it is both recognized by the LPR camera to depart from the upstream lane at time $t_m$ and depart from the downstream lane at time $t^{'}_m$. Under FIFO rules, the lane-based cumulative arrival index of the matched vehicle is equal to the lane-based cumulative departure index, i.e.,
\begin{align}
	A_{l,t_m}^*= D^* _{l, t_{m}^{'}},   
 \label{eq: Nm matching}
\end{align}
where $t_{m}^{'}$ is the timestamp of matched vehicles when passing the downstream approaching lane.

\subsection{Historical and Real-time ACR task}
\label{Historical and Real-time ACR task}
In the process of estimating the lane-based arrival curves, we utilize License Plate Recognition (LPR) data from both upstream and downstream locations, dividing the task into two distinct types: historical and real-time Arrival Curve Reconstruction (ACR). 
For historical lane-based ACR, there is an inherent time lag because once a vehicle is detected by the upstream LPR camera and then enters the target link, it takes a link travel time before being detected by the downstream LPR camera, as illustrated in Fig. \ref{fig: problem statement}(b). This time lag results in delayed updates to the historical ACR. 
In contrast, the real-time lane-based ACR aims to reconstruct the arrival curve instantaneously, regardless of this time lag. This makes real-time ACR more challenging than the historical ACR but particularly valuable for applications that require immediate traffic management responses.

\section{Methodology}
In this section, we formulate the lane-based arrival curve reconstruction problem based on lane choice learning in subsection \ref{subsec: problem transformation}. 
Then, a Bayesian arrival curve learner is designed to learn the lane choice patterns and uncertainties in subsection \ref{subsec: Arrival Curve Learning}.
Finally, the historical and real-time arrival curve reconstruction procedures are introduced in subsection \ref{subsec: Historical Arrival Curve Reconstruction}, respectively.

\subsection{Lane Choice Learning-Based Arrival Curve Reconstruction} \label{subsec: problem transformation}

The reconstruction of the lane-based arrival curve is equivalent to estimating the lane-based cumulative arrival index of any arriving vehicle. The existing approaches typically employed interpolation methods such as the Gaussian Process to infer the lane-based cumulative arrival indices of unmatched vehicles \cite{zhan2015lane, mo2017speed} based on cumulative indices of matched vehicles, which are extracted through vehicle license plate matching using LPR data upstream and downstream intersections, as detailed in subsection \ref{subsec: cumulative departure curves and matched vehicles}. While effective in historical arrival curve reconstruction, these methods depend on historical matched vehicles when applying to real-time reconstruction, weakening their reliability in real-time applications.

In this study, we design a lane choice learning-based arrival curve reconstruction, which enables the utilization of additional real-time link-based arrival observations from LPR data in real-time reconstruction. Specifically, the lane-based cumulative arrival index of any arriving vehicle can be expressed as
\begin{align}
	A_{l,t} =A_{l,t_m}^* + \Delta a_{l,m,t},
\label{eq: arrival index estimation}
\end{align}
where $A_{l,t_m}^*$ is the lane-based cumulative arrivals of the previous matched vehicle arriving at $t_m$, and $\Delta a_{l,m,t}$ is the lane-based arrival accumulations during the interval $\left( t_m,t \right]$. 
The lane-based arrival accumulations can be determined by utilizing the link-based arrival accumulations from LPR data and a lane choice proportion, i.e.,
\begin{align}
	\Delta a_{l,m,t}= \alpha_l \cdot \Delta s_{m,t}^{*},
\label{eq: turning ratio}
\end{align}
where $ \Delta s_{m,t}^{*}$ is the observed link-based arrival accumulations during the interval $\left( t_m,t \right]$, $\alpha_l$ is a lane choice proportion that distributes the link-based arrivals to the downstream lane $l$.
However, in practical scenarios, the link-based arrivals are partially observed due to the incomplete LPR systems monitoring of upstream approaching lanes, as demonstrated in Fig. \ref{fig: problem statement}(a). 
Therefore, instead of using the linear proportion, this study introduces a non-linear lane choice mapping function to capture the relationship between the partially observed link-based arrivals and lane-based arrivals, i.e.,  
\begin{align} 
	\Delta a_{l,m,t}= f\left(\Delta s_{m,t}^{*}\right),
\label{eq: arrival estimation}
\end{align}
where $f$ is the non-linear lane choice mapping function, $\Delta s_{m,t}^{*}$ is the partially observed link-based arrival accumulations, which come from multiple upstream lanes, i.e.,
\begin{align}
	\Delta s_{m,t}^{*} = \sum_{l'}{\Delta s_{l',m,t}^{*} }, 
\label{eq: link-based arrivals}
\end{align}
where $\Delta s_{l',m,t}^{}$ is the link-based arrival accumulations coming from upstream approaching lane $l'$.
According to Assumption 1, the link-based arrival accumulations from the upstream lane are equal to the upstream lane-based departure accumulations during $\left( t_m,t \right]$  i.e.,
\begin{align}
\Delta s_{l',m,t}^{*} = \Delta d_{l',m,t}^{*}  = D_{l',t}^{*}- D_{l',t_m}^{*},
\label{eq: link-based arrivals from upstream lanes}
\end{align}
where $D_{l',t}^{*}$ and $D_{l',t_m}^{*}$ is the observable cumulative departure index of upstream lane-based departure curves at time $t$ and $t_m$. 
The lane choice patterns of link-based arrivals from different upstream lanes are distinguishing. Therefore, to capture refined lane choices and avoid information loss, the Eq. \eqref{eq: arrival estimation} is rewritten as
\begin{align}
	\Delta a_{l,m,t}=f\left( \Delta \boldsymbol{s}_{l',m,t}^*\right),
\label{eq: arrival estimation with vector input}
\end{align}
where $ \Delta \boldsymbol{s}_{l',m,t}^*$ is a vector consisting of multiple observed upstream link-based arrival accumulations from different upstream lanes.

In order to better capture lane choice patterns, we introduce additional upstream signal phase information because different upstream signal phases could correspond to distinguishing lane choice patterns. For example, the upstream through phase probably contributes the most arrivals of downstream through lane. 
Therefore, we introduce two additional features, the time in the signal cycle and the estimation span, and reformulate the Eq. \eqref{eq: arrival estimation} as
\begin{align}
	\Delta a_{l,m,t}=f\left( \Delta \boldsymbol{s}_{l',m,t}^*, t_{m}^{c},\delta \right),
\label{eq: lane choice mapping function}
\end{align} 
where $t_{m}^{c}$ is the time in signal cycles of the referenced matched vehicles $m$, $\delta =t-t_m$ is the estimation span, i.e., the time difference between the current estimation and previous matched vehicles.
In addition to LPR data, the proposed framework is also compatible with the emerging CAV data, which can provide additional observations, e.g., link-based arrivals, about mid-block merging traffic, and right-turning traffic at intersections, which are usually not monitored by LPR cameras.

\subsection{Lane Choice Learning via Bayesian Deep Learning}
\label{subsec: Arrival Curve Learning}

In this subsection, we propose a Bayesian Arrival Curve Learner (BACL) to capture the lane choice patterns and quantify the lane choice uncertainties.
Specifically, the lane choice pattern learning using BACL is illustrated in subsection \ref{subsubsec: Arrival Curve Learner Design}. We specify how the lane choice uncertainties are quantified in subsection \ref{subsubsec: Bayesian deep learning}.

\subsubsection{Learning Lane Choice Patterns}\label{subsubsec: Arrival Curve Learner Design}
To learn lane choice patterns, we employ a Bayesian deep learning approach termed BACL to serve as the lane choice mapping function $f$, which is trained utilizing historical link-based and lane-based arrival observations.
Based on historical matched vehicles, The historical lane-based arrivals can be calculated as
\begin{align}
\Delta a^*_{l, m}= A_{l,t_{m+1}}^* - A_{l,t_m}^*,
\label{eq: lane-based arrival observations}
\end{align}
where $ A_{l,t_{m+1}}^* $ and $ A_{l,t_m}^*$ denote the cumulative arrival index of $m$ and $m+1$ matched vehicles. The historical link-based arrivals from upstream lane $l'$ can be calculated based on observed lane-based departure curves, i.e.,
\begin{align}
\Delta s_{l',m}^{*} = \Delta d_{l',m}^{*}  = D_{l',t_{m+1}}^{*}- D_{l',t_m}^{*},
\label{eq: link-based arrival observations}
\end{align}
where $D_{l',t_{m+1}}^{*}$ and $D_{l',t_m}^{*}$ is the cumulative departure index of upstream lane-based departure curves at time $t_{m+1}$ and $t_m$.
Given the historical lane-based and link-based arrival observations, the lane choice pattern learning using BACL is reformulated as follows
\begin{align}
	&\min \sum_{l=1}^L{\sum_{m=1}^{M_{l}-1}{F\left( \Delta a_{l,m}, \Delta a_{l,m}^{*} \right)}},\nonumber\\
	&~~s.t. ~ \Delta a_{l,m}=f ^{\omega} \left( \Delta \boldsymbol{s}_{l',m}^{*}, t_{m}^{c},\delta \right), 
 \label{eq: arrival curve learning}
\end{align}
where 
\begin{itemize}
    \item[$\Delta a_{l, m}$] is the estimated lane-based arrival accumulation during the time interval $\left( t_m, t_{m+1} \right]$, where $t_m$ and $t_{m+1}$ denote the arrival timestamp of $m^{th}$ and $(m+1)^{th}$ matched vehicles,
    \item[ $\Delta a^*_{l, m}$] is the observed upstream lane-based arrival accumulation during the time interval $\left( t_m, t_{m+1} \right]$, 
    \item[ $\Delta \boldsymbol{s}_{l',m}^*$] is a vector consisting of multiple observed link-based arrival accumulations from different upstream lanes $l'$, during the time interval $\left( t_m, t_{m+1} \right]$,  
    \item[ $F\left( \cdot \right)$] is a probabilistic loss function,
    \item[ $f ^{\omega} \left( \cdot \right)$] is the BACL-based lane choice mapping function,
    \item[$M_{l}$] is the number of matched vehicles in lane $l$,
    \item[$L$] is the number of lanes. 
\end{itemize}

\subsubsection{Learning Lane Choice Uncertainty}\label{subsubsec: Bayesian deep learning}
\begin{figure}[t]
    \begin{center}
    \includegraphics[width=3.5in]{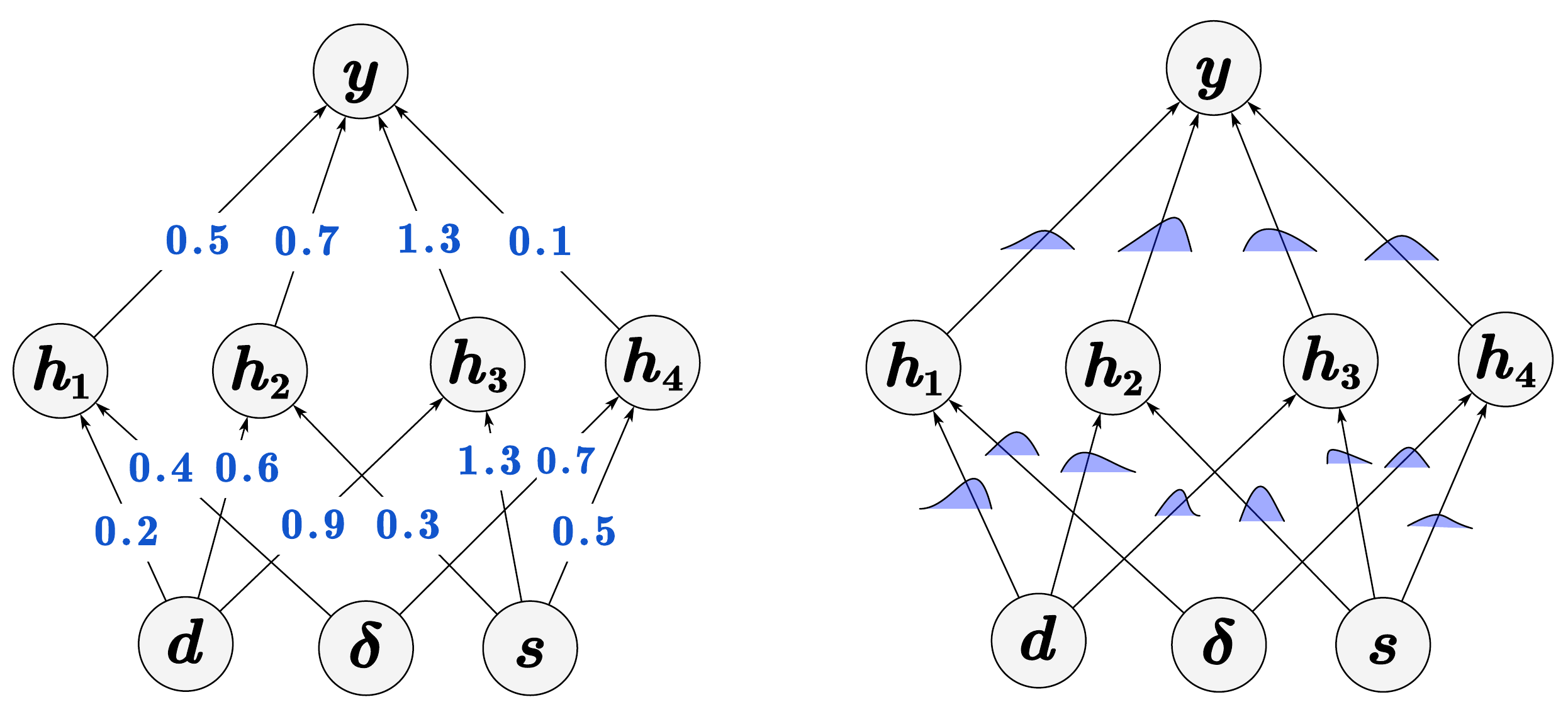}\\
	\caption{Left: Classical neural network, each weight has a fixed value. Right: Bayesian neural network, each weight is represented by a distribution.}
	\label{fig: BNN}
    \end{center}
\end{figure}

This subsection further specifies how lane choice uncertainties are characterized. To quantify the lane choice uncertainties, instead of using fixed parameter values, we put a prior distribution $p\left( \boldsymbol{\omega } \right)$ over the space of model parameters $\omega$ of the BACL-based lane choice mapping function $f ^{\omega} \left( \cdot \right)$, which represents our prior belief as to which parameters are likely to have mapped link-based arrivals to lane-based arrivals before we observe any matched vehicle data, as illustrated in Fig. \ref{fig: BNN}. In this study, we assume the Gaussian distribution as the prior and posterior distribution of model weights and the predictive distribution of estimated lane-based arrivals, given its computation simplicity, variational inference efficiency, and expressive capability.

For the sake of simplicity, we denote the link-based arrival accumulations, time in the signal cycle, and estimation span as model input, i.e., $\boldsymbol{x}=\left\{\Delta \boldsymbol{s}_{l',m}^{*},t_{m}^{c},\delta \right\} $, and the corresponding lane-based arrival accumulations as model output, i.e., $y= \Delta a_{l,m}$. The historical dataset can be expressed as $\mathbf{X}=\left\{ \boldsymbol{x}_1,...,\boldsymbol{x}_N \right\} $  and $\boldsymbol{y}=\left\{ y_1,...,y_{\mathrm{N}} \right\} $, and the posterior distribution over the space of parameters can be given by Bayes’s theorem
\begin{align}
	p\left( \boldsymbol{\omega }|\mathbf{X},\boldsymbol{y} \right) =\frac{p\left( \boldsymbol{y}|\mathbf{X},\boldsymbol{\omega } \right) p\left( \boldsymbol{\omega } \right)}{p\left( \boldsymbol{y}|\mathbf{X} \right)},
\label{eq: Bayes theorem}
\end{align}
where $p\left( \boldsymbol{y}|\boldsymbol{X},\boldsymbol{\omega } \right)$ denotes the likelihood distribution by which the model inputs, e.g., link-based arrival accumulations, generate the lane-based arrivals given parameter setting $\boldsymbol{\omega}$.
Then, we can estimate lane-based arrival accumulation given new model inputs by integrating all likelihood generated from the posterior distribution, i.e.,
\begin{align}
	p\left( \hat{y}| \hat{\boldsymbol{x}},\mathbf{X},\boldsymbol{y} \right) =\int{p\left( \hat{y}| \hat{\boldsymbol{x}},\boldsymbol{\omega } \right)}p\left( \boldsymbol{\omega }|\mathbf{X},\boldsymbol{y} \right) \mathrm{d}\boldsymbol{\omega },
\label{eq: Bayes prediction}
\end{align}
where $\hat{\boldsymbol{x}}$ denotes the new model inputs, $\hat{y}$ represents the estimated lane-based arrival accumulation, and $p\left( \hat{y}| \hat{\boldsymbol{x}},\boldsymbol{\omega } \right)$ is the likelihood.

\subsubsection{Variational Inference and Parameter Update}

\begin{figure}[t]
    \begin{center}
    \includegraphics[width=2.7in]{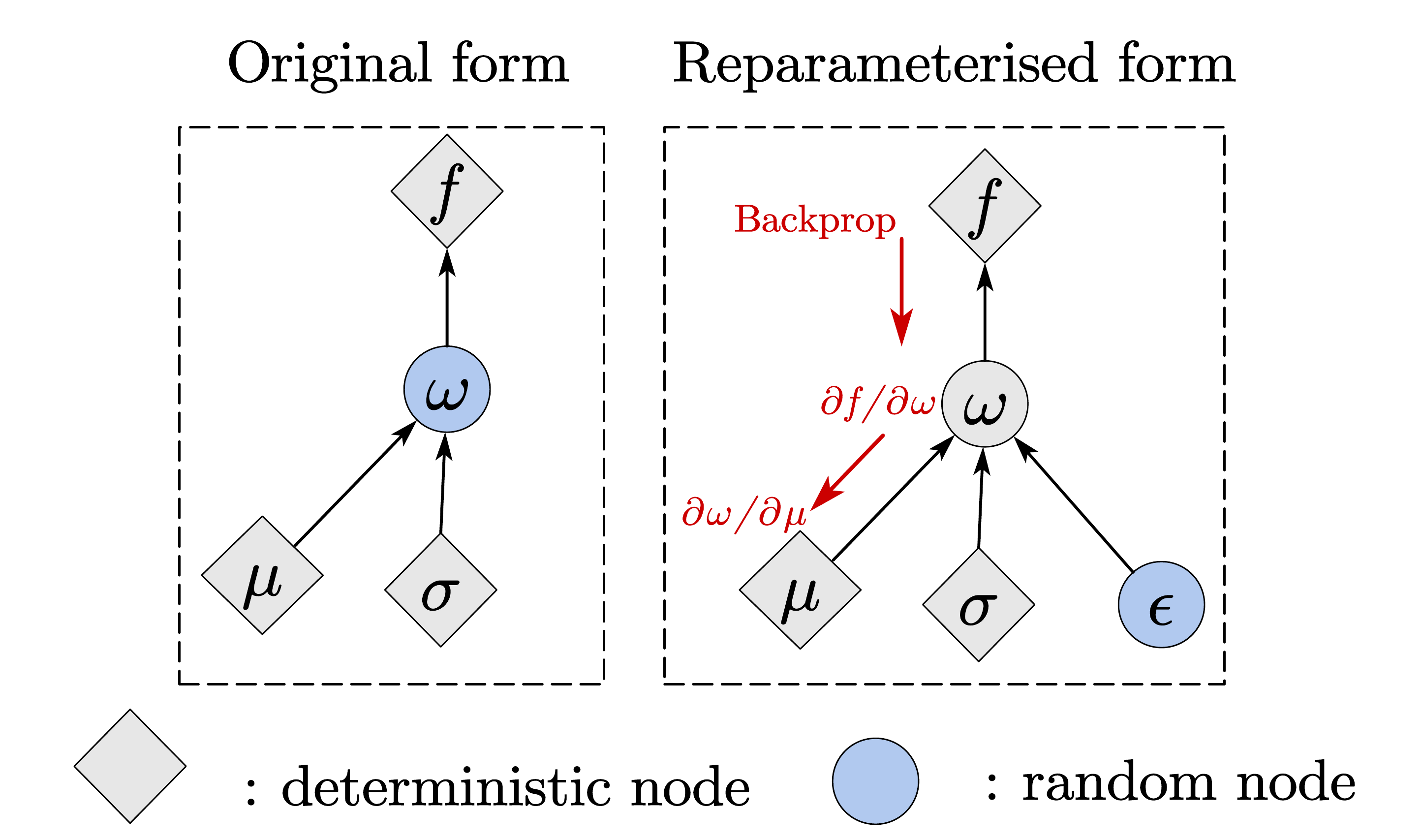}\\
	\caption{Illustration of the reparameterization trick.} 
	\label{fig: reparameterise}
    \end{center}
\end{figure}

It is intractable to solve the posterior distribution of weights $p\left( \boldsymbol{\omega }|\mathbf{X},\boldsymbol{y} \right) $ directly using Bayes theorem in Eq. \eqref{eq: Bayes theorem}. Therefore, we use a variational weight distribution $q_{\theta}\left( \omega \right)$ to approximate the actual posterior. The parameters of the variational weight distribution can be estimated by minimizing the KL divergence between the variational distribution $q_{\theta}\left( \omega \right) $ and the actual posterior $p\left( \boldsymbol{\omega }|\mathbf{X},\boldsymbol{y} \right) $, i.e.,
\begin{align}
	\hat{\theta}&=\underset{\theta}{\mathrm{arg}\min} ~\mathrm{KL}\left( q_{\theta}\left( \boldsymbol{\omega } \right) ||p\left( \boldsymbol{\omega }|\mathbf{X},\boldsymbol{y} \right) \right) \nonumber\\
	&=\underset{\theta}{\mathrm{arg}\min}\int{q_{\theta}\left( \boldsymbol{\omega } \right)}\log \frac{q_{\theta}\left( \boldsymbol{\omega } \right)}{p\left( \boldsymbol{\omega } \right) p\left( \boldsymbol{y}|\mathbf{X},\boldsymbol{\omega } \right)}d\boldsymbol{\omega } \nonumber\\
	&=\underset{\theta}{\mathrm{arg}\min} ~\mathrm{KL}\left( q_{\theta}\left( \boldsymbol{\omega } \right) ||p\left( \boldsymbol{\omega } \right) \right) -\mathbb{E} _{q_{\theta}\left( \boldsymbol{\omega } \right)}\left[ \log p\left( \boldsymbol{y}|\mathbf{X},\boldsymbol{\omega } \right) \right]
	\label{eq: KL divergence},
\end{align}
which is equivalent to minimizing negative Evidence Lower Bound (ELBO)
\begin{align}
  {\mathrm{KL}\left[ q_{\theta}\left( \boldsymbol{\omega } \right) ||p\left( \boldsymbol{\omega } \right) \right]}-\sum_{i=1}^N{\int{q_{\theta}\left( \boldsymbol{\omega } \right)}\log p\left( y_i|\mathrm{f}^{\boldsymbol{\omega }}\left( \boldsymbol{x}_i \right) \right) \mathrm{d}\boldsymbol{\omega }}\boldsymbol{x}_{\mathrm{i}},
	\label{eq: ELBO}
\end{align}
where $\boldsymbol{x}_i$ and $y_i$ denote the $i$ sample in the training dataset, and $N$ represents the number of samples. 
In forward arrival estimation, directly sampling the weights from the variational posterior distribution  $q_{\theta}\left( \omega \right) $ will lose the corresponding gradients that are essential in the backward parameter estimation process. 
Therefore, to keep the gradient and reserve the uncertainties of arrival estimation, we use the Gaussian reparameterization trick. Specifically, the sampling process is removed from the model structure, and the posterior weight $\omega$ is generated using a parameter-free noise distribution $\epsilon \sim \mathcal{N} \left( 0,I \right) $, as shown in Fig. \ref{fig: reparameterise}. The corresponding variational posterior parameters $\theta =\left( \mu ,\rho \right) $ is expressed as
\begin{align}
	\boldsymbol{\omega }=t\left( \theta ,\epsilon \right) =\mu +\log \left( 1+\exp \left( \rho \right) \right) \circ \epsilon, 
\label{eq: reparameterise}
\end{align}
where $\circ$ is point-wise multiplication. According to the Lemma 1 in \citep{blundell2015weight}, the gradients with respect to the mean value and standard deviation parameter of the variational distribution are given as
\begin{align}
&\varDelta _{\mu}=\frac{\partial f\left( \omega ,\theta \right)}{\partial \omega}+\frac{\partial f\left( \omega ,\theta \right)}{\partial \mu},
\\
&\varDelta _{\rho}=\frac{\partial f\left( \omega ,\theta \right)}{\partial \omega}\frac{\epsilon}{1+\exp \left( -\rho \right)}+\frac{\partial f\left( \omega ,\theta \right)}{\partial \rho},
\label{eq: gradient mu and sigma}
\end{align}
where $f\left( \omega ,\theta \right) =\log q_{\theta}\left( w \right) -\log p\left( w \right) p\left( \boldsymbol{y}|\mathbf{X},w \right)$. The parameters of the variational distribution are updated as
\begin{align}
	\mu \gets \mu -\alpha \varDelta _{\mu},
	\\
	\rho \gets \rho -\alpha \varDelta _{\rho},
\label{eq: update mu and sigma}
\end{align}
where $\alpha$ denotes a learning rate.

\subsubsection{Arrival Accumulation Estimation and Uncertainty Quantification}
Given the estimated variational posterior distribution, the mean value and variance of estimated arrival accumulation are derived by generating  weight samples from the variational distribution, i.e.,
\begin{align}
    \mathbb{E} \left[ \hat{y} \right] =\int{p\left( \hat{y}|\boldsymbol{\hat{x}},\boldsymbol{\omega } \right)}q_{\theta}\left( \boldsymbol{\omega } \right) d\boldsymbol{\omega }\approx \frac{1}{M}\sum\nolimits_{m=1}^M{f^{\hat{\omega}_m}\left( \boldsymbol{\hat{x}} \right)},
\label{eq: mean value of arrival estimations}
\end{align}
\begin{align}
    \mathrm{Var}\left[ \hat{y} \right] &= \underset{\mathrm{epistemic} ~\mathrm{lane}~ \mathrm{choice} ~\mathrm{uncertainty}}{\underbrace{\frac{1}{M}\sum\nolimits_{m=1}^M{\left[ f^{\hat{\omega}_m}\left( \boldsymbol{\hat{x}} \right) \right] ^2}-\mathbb{E} \left[ \hat{y} \right] ^2}} \nonumber \\
    &+\underset{\mathrm{aleatoric} ~\mathrm{arrival}~ \mathrm{uncertainty}}{\underbrace{\frac{1}{M}\sum\nolimits_{m=1}^M{\sigma _n\left( \boldsymbol{\hat{x}} \right)}}},
\label{eq: variance of arrival estimations}
\end{align}
where $ f^{\hat{\omega}_m}\left( \boldsymbol{\hat{x}} \right)$ is the  estimated arrivals of BACL model for the input $\hat{\boldsymbol{x}}$ using the generated weight $\hat{\omega}_m$ sample from the variational posterior distribution $q_{\theta}\left( \omega \right)$, $M$ is the number of samples,  $\hat{y}=\Delta a_{l,m,t}$, $\hat{\boldsymbol{x}} = \left( \Delta \boldsymbol{s}_{l',m,t}^*, t_{m}^{c},\delta \right)$.

\subsection{Historical and Real-time Arrival Curve Reconstruction}
\label{subsec: Historical Arrival Curve Reconstruction}

The historical Arrival Curve Reconstruction (ACR) aims to estimate the lane-based cumulative arrival index of historical unmatched vehicles, while the real-time ACR seeks to estimate the lane-based cumulative arrival index of real-time arriving vehicles.

\subsubsection{Historical ACR}
The first step in historical ACR is to determine the time range of reconstructed curves, which is defined as the range between the arriving time of two adjacent matched vehicles, i.e.,
\begin{align}
	T_m=\left( t_m,t_{m+1} \right] , m=1,2,...,M_l-1,
\label{eq: time interval}
\end{align}
where $T_m$ is the $m$ reconstructed time range of the arrival curves.
Based on Eq. \eqref{eq: mean value of arrival estimations}, we can obtain the mean estimation of arrival accumulations $\Delta \hat{a}_{l,m,t}$.
The two adjacent matched vehicles provide boundary constraints that limit the up value of estimated arrival accumulations. 
The estimated lane-based arrival accumulations within the range $T_m$ must equal the observed one, which is calculated in Eq. \eqref{eq: lane-based arrival observations}.
For each reconstructed area $T_m$, we introduced a modified factor to incorporate the boundary constraint into the estimation of arrival accumulations, i.e.,
\begin{align}
	\lambda _m=\frac{\Delta a_{l,m}^*}{\Delta \hat{a}_{l,m,t}},
\label{eq: scaling parameter}
\end{align}
where $\Delta \hat{a}_{l,m,t}$ is the estimated arrival accumulation.
The Eq. \eqref{eq: arrival index estimation} can be written as
\begin{align}
	A_{l, t} = A_{l,t_m }^{*} + \lambda_m \cdot \Delta\hat{a}_{l,m,t}.
\label{eq: ACR}
\end{align}
The pseudocode of historical ACR is presented in Algorithm \ref{alg: historical ACR}.

\begin{algorithm}[t]
	\caption{Historical Lane-based Arrival Curve Reconstruction}
	\begin{algorithmic}[1]
		\renewcommand{\algorithmicrequire}{\textbf{Input:}  The historical LPR data set, signal timing;} 
		\renewcommand{\algorithmicensure}{\textbf{Output:} The historical arrival curve $A_l\left( t \right), t< t_M$;}
        \REQUIRE 
        \ENSURE 
        \STATE Extract lane-based cumulative arrival index of matched vehicle observations $A_{l,m}^*, m=1,2,...,M_l -1 $ using Eq. \eqref{eq: Nm matching} 
        
        \FOR{$m$=1:$M_l$-1}
        \STATE Select the $m$ matched vehicle as a reference;
        \FOR{$t$ in range $\left( t_{m},t_{m+1} \right] $}
        \STATE Calculate the reconstructed time span $\delta =t-t_m$;
        \STATE Calculate the link-based arrival accumulations merged from different upstream lanes via Eq. \eqref{eq: link-based arrivals from upstream lanes};
        \STATE Estimate the mean value and variance of the estimated lane-based arrival accumulations $\Delta \hat{a}_{l,m,t}$ via Eq. \eqref{eq: lane choice mapping function}, Eq. \eqref{eq: mean value of arrival estimations}, and Eq. \eqref{eq: variance of arrival estimations};
        \STATE Modify the mean estimated arrival accumulation and estimate the lane-based cumulative arrival index of unmatched vehicles via Eq. \eqref{eq: ACR};
        \ENDFOR
        \ENDFOR
	\end{algorithmic}
	\label{alg: historical ACR}
\end{algorithm}

\subsubsection{Real-time ACR}

In real-time reconstruction, the real-time link-based arrival accumulations after the newest matched vehicle $M$ can be obtained from observed lane-based departure curves using real-time LPR data
\begin{align}
\Delta s^*_{l',M,t} = D^*_{l',t} -D^*_{l', t_M},
\label{eq: real-time link-based arrivals}
\end{align}
where $D^*_{l',t} $ and $D^*_{l', t_M}$ are the observed cumulative departure indices of lane-based departure curves at time $t$ and $t_M$.
Based on the real-time link-based arrivals, we can estimate the mean and variance of real-time lane-based arrival accumulations using the trained BACL via Eq. \eqref{eq: mean value of arrival estimations} and Eq. \eqref{eq: variance of arrival estimations}, and the real-time arrival curve reconstruction can be expressed as
\begin{align}
	A_{l,t}=A_{l,t_M}^{*}+\Delta \hat{a}_{l,M,t},
\label{eq: RACR}
\end{align}
where $A_{l,t_M}^{*}$ is the lane-based cumulative arrival index of the newest matched vehicle, $ \Delta \hat{a}_{l,M,t}$ is the mean estimation of real-time lane-based arrival accumulation using trained BACL. The pseudocode of real-time ACR using real-time LPR data is presented in Algorithm \ref{alg: real-time ACR}.

For practical applications, the real-time cumulative arrival curve can be applied to real-time lane-based vehicle count estimation. Specifically, given the estimated real-time lane-based arrival curve $A_l\left( t \right)$, the real-time lane-based vehicle count $G_l({t})$ can be determined as
\begin{align}
	G_l\left( {t} \right) =A_l\left( {t} \right) -D_l^*\left( {t} \right),
\label{eq: lane storage}
\end{align}
where $D_l^*\left({t} \right)$ is the lane-based downstream departure curve. We present the estimation examples of real-time lane-based vehicle count estimation in subsection \ref{Real-time ACR results}.

\begin{algorithm}[t]
	\caption{Real-time Lane-based Arrival Curve Reconstruction}
	\begin{algorithmic}[1]
		\renewcommand{\algorithmicrequire}{\textbf{Input:}  The real-time LPR data set, signal timing;} 
		\renewcommand{\algorithmicensure}{\textbf{Output:} The real-time arrival curve $A_l\left( t \right), t> t_M$;}
        \REQUIRE 
        \ENSURE 
        \STATE Extract the lane-based cumulative arrival index of the newest matched vehicle $A_{l,M}^*$ using Eq. \eqref{eq: Nm matching}; 
        \STATE Select the newest matched vehicle $M$ as a reference;
        \WHILE{No new matched vehicle update}
        \STATE Select the reconstructed timestamp $t = t_{current}$; 
        \STATE Calculate the reconstructed time span $\delta =t-t_M$;
        \STATE Calculate the link-based arrival accumulations merged from different upstream lanes via Eq. \eqref{eq: real-time link-based arrivals};
        \STATE Estimate the mean value and variance of the estimated lane-based arrival accumulations $\Delta \hat{a}_{l,M,t}$ via Eq. \eqref{eq: lane choice mapping function}, Eq. \eqref{eq: mean value of arrival estimations}, and Eq. \eqref{eq: variance of arrival estimations};
        \STATE Obtain the real-time lane-based arrival curve $A_l(t)$ via Eq. \eqref{eq: RACR};
        \ENDWHILE
	\end{algorithmic}
	\label{alg: real-time ACR}
\end{algorithm}

\section{Experiments}

In this section, we evaluate our proposed method on a real-world LPR dataset. We introduce the test site, baseline methods, and evaluation metrics in subsection \ref{experimental settings}. The historical arrival curve reconstruction model performances are evaluated in subsection \ref{Historical ACR results under different matching rates}. The real-time estimation performances of lane-based arrival curves and vehicle counts are assessed in subsection \ref{Real-time ACR results}. Finally, we inspect the model performance using different utilization levels of LPR data in subsection \ref{Ablation study}.

\subsection{Experimental settings}\label{experimental settings}

\subsubsection{Test Site Description}
\label{test site}

\begin{figure}[t]
    \begin{center}
    \includegraphics[width=3.5in]{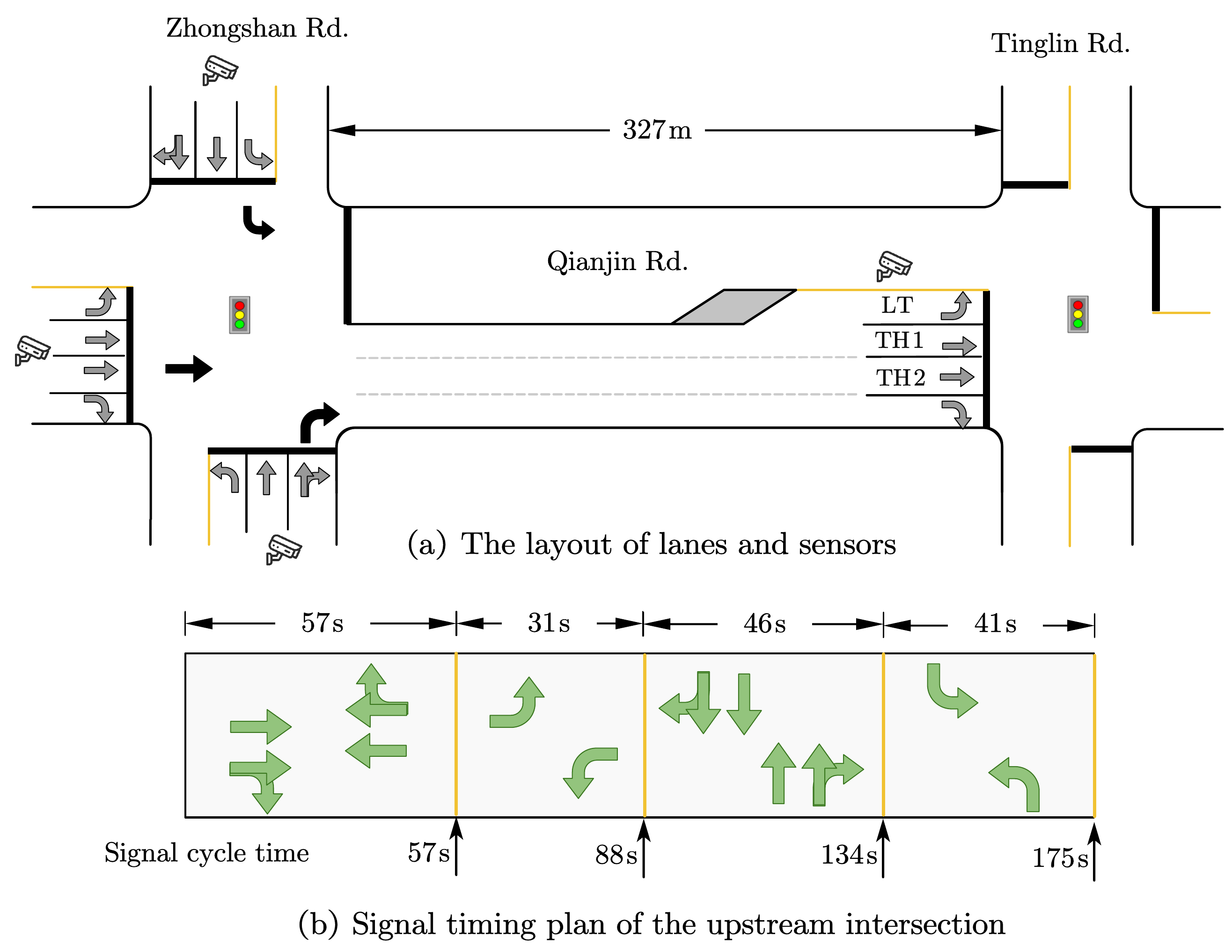}\\
	\caption{Visualization of the test site layout and signal timing.}
	\label{fig: Test site}
    \end{center}
\end{figure}

In field experiments, we select two adjacent intersections and the connecting link on Qianjing Road in Kunshan, Jiangsu Province, China as our test sites. 
As shown in Fig. \ref{fig: Test site}, the LPR data are collected at the upstream and downstream intersections from June 4th to June 8th, 2018, spanning a total of five days. The evaluation includes three traffic scenarios, each lasting 15 minutes: the morning peak (AP) from 8:15 to 8:30 AM, the off-peak (OP) from 2:15 to 2:30 PM, and the evening peak (PP) from 5:15 to 5:30 PM.
Three of four downstream lanes are selected as test lanes: the left-turn lane (LT), through lane 1 (TH1), and lane 2 (TH2).
Three of the five days' data are used to learn the arrival curves, while the data from the remaining two days are served for evaluating the reconstruction of arrival curves.
The LPR sensors at the test site did not work well and the matching rate of LPR data is 61.6$\%$ on average. Additional video data are also collected to fix unrecognized vehicle license plates, improving the matching rate to 97.8$\%$. This corrected dataset is used to provide ground truth for model validation.

\subsubsection{Benchmark Method}
We compared the proposed Bayesian method with the State-Of-The-Art (SOTA) probabilistic and deterministic methods.
\begin{itemize}
    \item \textbf{AF-GP} (Arrival Function-based Gaussian Process \cite{zhan2015lane,mo2017speed}):  A customized Gaussian process for arrival curve reconstruction, in which a customized mean function is calibrated by two-stage cyclic arrival rates to capture arrival patterns and a covariance function is developed to characterize arrival uncertainties. The arrival curves are reconstructed by inferring the cumulative arrival indices of unmatched vehicles using the AF-GP model based on matched vehicles.

    \item \textbf{LC-NN} (Lane Choice-based Feedforward Neural Network): A plain neural network that reconstructs the arrivals curves based on lane choice learning, using matched vehicles and real-time link-base arrivals.
    The LC-NN is used for comparison to demonstrate the superior performance of the Bayesian approach in modeling lane choice uncertainty.
    
\end{itemize}

\subsubsection{Evaluation Metrics}

To assess the arrival curve reconstruction performance, we employ a deterministic metric Root Mean Square Error (RMSE) and a probabilistic metric Continuous Ranked Probability Score (CRPS). The metric RMSE is used to evaluate the mean estimation of cumulative arrivals
 \begin{align}
    &\mathrm{RMSE}=\sqrt{\frac{1}{n}\sum\nolimits_{i=1}^n{\begin{array}{c}
        \left( a_i^*-{a}_i \right) ^2\\
	\end{array}}},\nonumber\\
 \label{eq: RMSE and MAPE}
\end{align}
where $a_i^*$ and $a_i$ represent the actual value and estimated mean value of the $i^{th}$ unobserved cumulative arrivals, respectively. 

In addition to the deterministic RMSE, the CRPS is employed to serve as a probabilistic evaluation of the estimated arrival distribution, reflecting both the accuracy of the mean estimation and the appropriateness of the uncertainty quantification
\begin{align}
\mathrm{CRPS} =\int\limits_{-\infty}^{\infty}{\left( F\left( a \right) -H\left( a-a^*\right) \right) ^2da},
\end{align}
where $F(a)$ is the cumulative distribution function (CDF) of the estimated distribution, $a^*$ is the observed value, $H$ is the Heaviside step function, i.e., 0 for $a<a^*$ and 1 for $a\geqslant a^*$. A lower CRPS value indicates that the distribution estimate is closer to the actual observed value.

\subsection{Historical Arrival Curve Reconstruction Results }
\label{Historical ACR results under different matching rates}

To evaluate the model performance of historical arrival curve reconstruction (ACR) from both deterministic and probabilistic perspectives, we summarize the average RMSEs and CRPSs (including standard deviations) across various LPR data matching rates in Tab. \ref{tab: Average RMSEs} and Tab. \ref{tab: Average CRPSs}. The metrics are calculated using testing data spanning nine scenarios (3 testing lanes $\times$ 3 testing periods).
The proposed BACL model consistently outperforms the baseline methods across diverse matching rates. 
Tab. \ref{tab: Average RMSEs} shows that the RMSEs for interpolation-based AF-GP remain relatively stable when matching rates (MRs) exceed $40 \%$, but they increase significantly, with larger standard deviations, at MRs below $ 20 \%$.

In contrast, the lane choice-based neural network (LC-NN) and proposed BACL, which utilize link-based arrival information, display stable RMSEs with small variance even at low MRs. This stability suggests the effectiveness of the lane choice-based learning mechanism in reconstructing arrival curves.
By comparing the BACL with LC-NN, it can be seen that while the LC-NN performs competitively with the BACL at high matching rates, the LC-NN's RMSE increases dramatically in low matching rates scenarios where the uncertainties of lane choices are even higher, highlighting the value of proposed Bayesian deep learning approach.

By comparing the probabilistic CRPSs of the baseline and proposed method in Tab. \ref{tab: Average CRPSs}, we can observe that the CRPSs of AF-GP increase sharply in low MRs, while those for BACL remain robust with smaller variance. This consistency indicates that the estimated arrival distributions from BACL more closely align with the actual values, demonstrating the superiority of the proposed Bayesian deep learning approach in capturing lane choice and arrival uncertainties.

\begin{table}[t]
  \centering
  \caption{Average RMSEs with standard deviations of historical ACR under various matching rates.}
    \begin{tabular}{cccc}
    \toprule
    \multirow{2}[4]{*}{\textbf{Matching Rate}} & \multicolumn{3}{c}{\textbf{Model}} \\
\cmidrule{2-4}          & \textbf{AF-GP} & \textbf{LC-NN} & \textbf{BACL} \\
    \midrule
    10    & 5.903±2.415 & 2.201±0.281 & \textbf{1.607±0.200} \\
    20    & 3.485±1.971 & 1.830±0.175 & \textbf{1.359±0.106} \\
    30    & 2.907±1.764 & 1.445±0.071 & \textbf{1.224±0.170} \\
    40    & 2.203±0.899 & 1.374±0.157 & \textbf{0.926±0.439} \\
    50    & 1.886±0.901 & 1.136±0.053 & \textbf{0.736±0.165} \\
    60    & 1.552±0.419 & 1.063±0.048 & \textbf{0.734±0.152} \\
    70    & 1.422±0.529 & 0.901±0.035 & \textbf{0.615±0.216} \\
    80    & 1.138±0.280 & 0.807±0.028 & \textbf{0.531±0.135} \\
    90    & 1.109±0.340 & 0.681±0.017 & \textbf{0.504±0.152} \\
    \bottomrule
    \end{tabular}%
  \label{tab: Average RMSEs}%
\end{table}%

\begin{table}[t]
  \centering
  \caption{Average CRPSs with standard deviations of historical ACR under various matching rates.}
    \begin{tabular}{ccc}
    \toprule
    \multicolumn{1}{c}{\multirow{2}[4]{*}{\textbf{Matching Rate}}} & \multicolumn{2}{c}{\textbf{Model}} \\
\cmidrule{2-3}          & \textbf{ACR-GP} & \textbf{BACL} \\
    \midrule
    10    & 5.054±0.997 & \textbf{0.866±0.101} \\
    20    & 2.670±0.767 & \textbf{0.624±0.045} \\
    30    & 1.830±0.292 & \textbf{0.609±0.078} \\
    40    & 1.289±0.155 & \textbf{0.503±0.077} \\
    50    & 0.992±0.173 & \textbf{0.500±0.053} \\
    60    & 0.876±0.113 & \textbf{0.423±0.053} \\
    70    & 0.754±0.101 & \textbf{0.357±0.053} \\
    80    & 0.692±0.049 & \textbf{0.354±0.059} \\
    90    & 0.632±0.048 & \textbf{0.336±0.061} \\
    \bottomrule
    \end{tabular}%
  \label{tab: Average CRPSs}%
\end{table}%

\begin{figure*}[t]
  \begin{center}
  \includegraphics[width=6in]{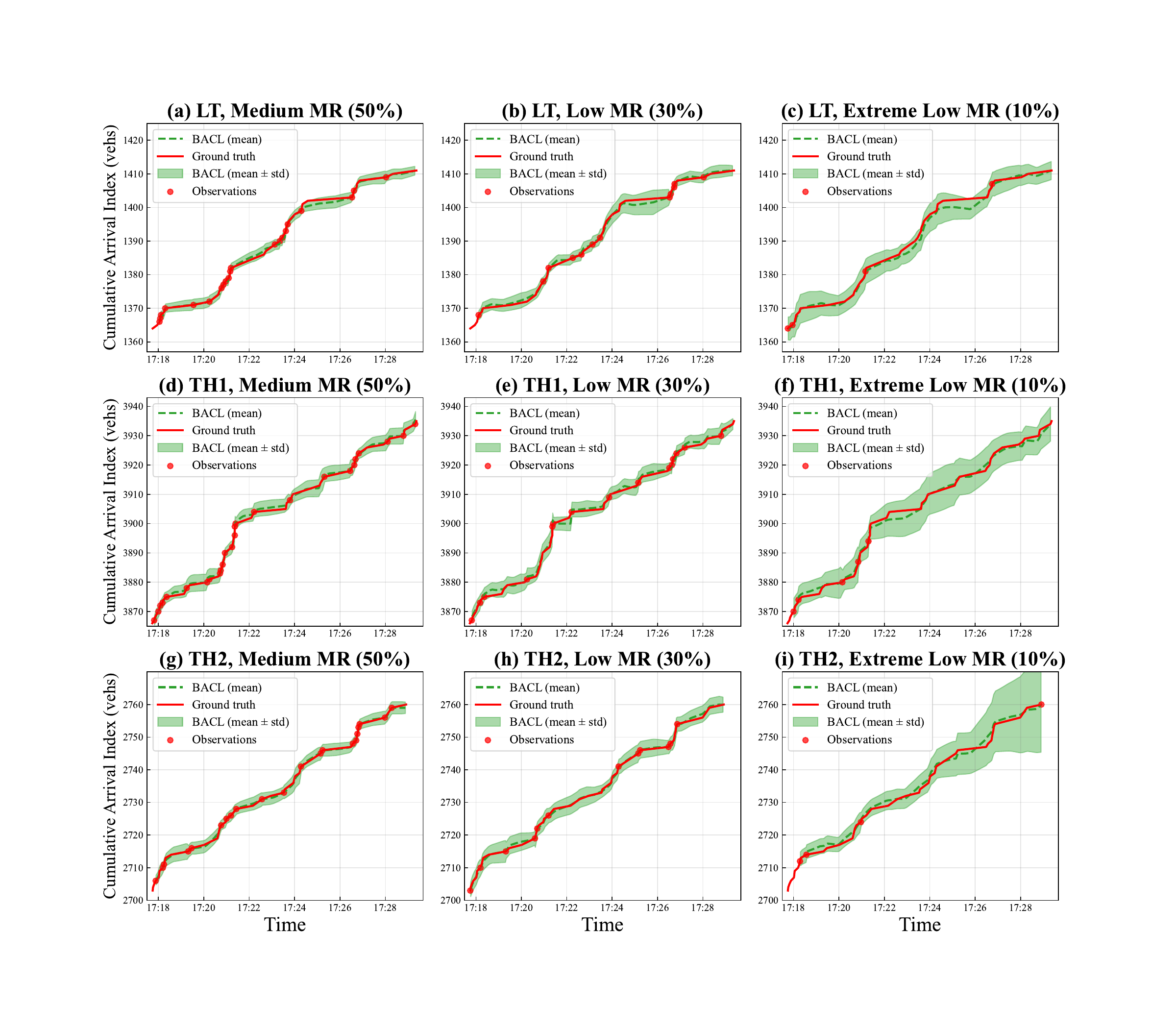}\\
\caption{Historical arrival curve reconstruction examples with mean estimates and uncertainty characterization.}
\label{fig: Performance examples}
  \end{center}
\end{figure*}

To vividly demonstrate the ACR performance, we visualize nine examples from three downstream lanes during PM peak hours across three matching rates (MRs) of 50$\%$, 30$\%$, and 10$\%$, shown in Fig. \ref{fig: Performance examples}. The green dashed lines and filled areas represent the BACL’s estimated arrival mean values and uncertainties, respectively.
It can be seen that the BACL consistently offers accurate and reliable reconstruction results with reasonable uncertainty characterization.
In the medium and low MR scenarios, e.g., Fig. \ref{fig: Performance examples}(a), (d), (g), the BACL's mean estimates closely align with the ground truth arrival curves. With relatively sufficient observations, the associated uncertainty estimates are small and trend consistently with the mean estimates.
In scenarios with extremely low MRs, e.g., Fig. \ref{fig: Performance examples}(c), (f), (i), the mean estimates may slightly diverge from the ground truth. However, the accompanying uncertainty (primarily lane choice uncertainty) estimates enhance reliability, ensuring the results remain robust even with limited data.

\begin{figure}[t]
    \begin{center}
    \includegraphics[width=3.5in]{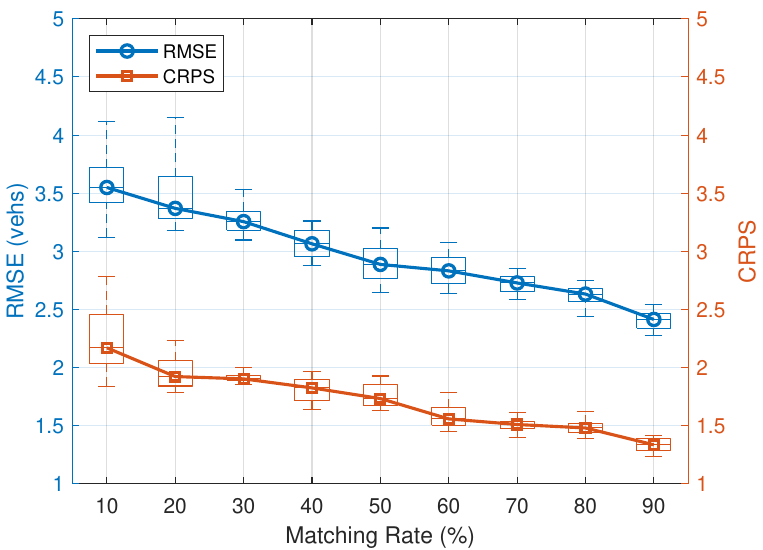}\\
	\caption{The RMSE and CRPS distribution of real-time arrival curve reconstruction under various matching rates.}
	\label{fig: ACP performance-prediction span}
    \end{center}
\end{figure}

\begin{figure*}[t]
  \begin{center}
  \includegraphics[width=5.5in]{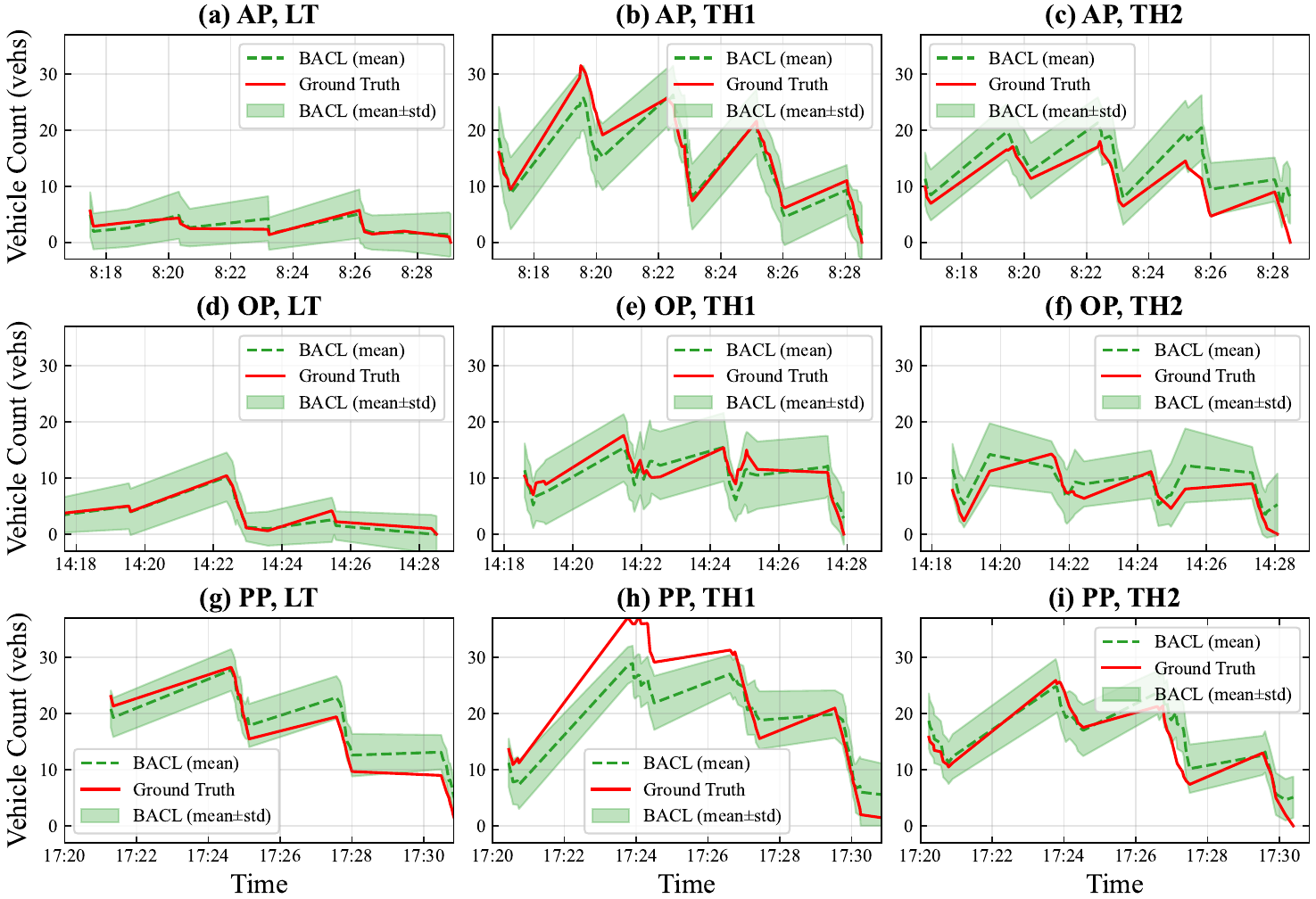}\\
\caption{Real-time lane-based vehicle count estimation examples.}
\label{fig: Real-time lane storage}
  \end{center}
\end{figure*}

\subsection{Real-time Arrival Curve Reconstruction Results}
\label{Real-time ACR results}

To assess the real-time reconstruction performance, we display the distributions of RMSE and CRPS under various matching rates, as shown in Fig. \ref{fig: ACP performance-prediction span}. For each matching rate, experiments are repeated 20 times. The links connecting the boxplot centers represent the median values of RMSEs and CRPSs. 
Compared to the historical reconstruction results in Tab. \ref{tab: Average RMSEs} and Tab. \ref{tab: Average CRPSs}, the RMSEs and CRPSs for real-time reconstruction are even higher. This increase is attributed to the historical reconstructions benefiting from bilateral boundary constraints, e.g., the adjacent matched vehicles illustrated in Fig. \ref{fig: problem statement}(b). Also, the estimation span is usually smaller than the update time of matching vehicles, e.g., 31s$\sim$141s.
In contrast, the real-time reconstruction faces more stringent challenges with only unilateral boundary constraints (e.g., the newest matched vehicle), and the estimation span is typically larger than the vehicle link travel time, e.g., 265s$\sim$437s.
Despite numerical differences, the RMSE and CRPS metrics exhibit consistent trends across varying matching rates.
As the matching rates increase, the values of both metrics improve, showing decreased values and smaller variances. This improvement is due to higher matching rates bringing the estimation span closer to the vehicle link travel time, therefore reducing the lane choice uncertainties associated with link-based arrivals.
Compared to the deterministic RMSE, the variances of probabilistic CRPSs are even lower at the same matching rates, indicating the reliability of our approach in quantifying lane choice uncertainties.

To further show the real-time arrival reconstruction performance,  we visualize nine real-time lane-based vehicle count estimation examples from three lanes during three testing periods under $50\%$ matching rates, as shown in Fig. \ref{fig: Real-time lane storage}. 
Overall, the BACL provides accurate lane-based vehicle count estimates with reliable uncertainties, indicating the superiority of the proposed BACL in estimating real-time operational metrics.

\begin{table}[t]
  \centering
  \caption{Significance of input features}
    \begin{tabular}{cccc}
    \toprule
    \multicolumn{1}{l}{\textbf{Matching Rate}} & \multicolumn{3}{c}{\textbf{Model}} \\
\cmidrule{2-4}          & \textbf{w/o link arrivals} & \textbf{w/o lane vector} & \textbf{BACL} \\
    \midrule
    10    & 2.582±0.094 & \textbf{1.479± 0.264} & 1.607±0.200 \\
    20    & 1.884±0.068 & \textbf{1.138±0.152} & 1.359±0.106 \\
    30    & 1.622±0.188 & \textbf{1.018±0.279} & 1.224±0.170 \\
    40    & 1.189±0.135 & 0.940±0.248 & \textbf{0.926±0.439} \\
    50    & 1.018±0.137 & 0.769±0.195 & \textbf{0.736±0.165} \\
    60    & 1.020±0.107 & 0.790±0.149 & \textbf{0.734±0.152} \\
    70    & 0.983±0.132 & 0.655±0.200 & \textbf{0.615±0.216} \\
    80    & 0.877±0.120 & 0.614±0.351 & \textbf{0.531±0.135} \\
    90    & 0.878±0.136 & 0.541±0.215 & \textbf{0.504±0.152} \\
    \bottomrule
    \end{tabular}%
  \label{tab: Significance of input features}%
\end{table}%

\subsection{Significance of different input features} \label{Ablation study}
To examine the significance of input features, two additional BACL models are configured using different levels of LPR data features in Tab. \ref{tab: Significance of input features}.
First, in the w/o link arrivals setting, the link-based arrivals input is removed in the BACL model, making it challenging to learn the lane choices from link-based arrivals to lane-based arrivals.
As a result, the model degrades sharply across various matching rates, implying the importance of incorporating link-based arrival data.
Second, the w/o lane vector setting uses aggregated link-based arrivals instead of a detailed vector input that contains upstream lane information. Interestingly, in scenarios with low MRs, this simplification results in lower RMSEs. The reduced complexity of this model variant requires fewer parameters, making it more suitable for conditions with limited data.

\section{Conclusion}\label{conclusions}

In this study, a Bayesian deep learning approach is developed to reconstruct real-time lane-based traffic arrival curves using LPR data. Targeting real-time reconstruction, a lane choice learning process is designed to effectively capture the relationship between link-based arrivals and lane-specific arrivals. Moreover, the lane choice uncertainties are characterized using Bayesian parameter inference techniques, minimizing the arrival curve reconstruction uncertainties. 
Extensive experiment results demonstrate the superiority and necessity of lane choice modeling in reconstructing historical and real-time lane-based arrival curves.

There are several future directions to advance this study.
First, the same matching rates of LPR data are used in the arrival curve learning and reconstruction stages. It is practically meaningful to test the model performance with different matching rates at the two stages because the matching rate in reality is not stationary due to equipment aging or maintenance.
Second, it is possible to extend the proposed link-level model to a network-scale traffic arrival acquisition model, in which the proposed model can serve as a sub-module to obtain link-level traffic demand.

\ifCLASSOPTIONcaptionsoff
  \newpage
\fi




\bibliographystyle{IEEEtranN}
\bibliography{ref}

\begin{thebibliography}{32}
\providecommand{\natexlab}[1]{#1}
\providecommand{\url}[1]{#1}
\csname url@samestyle\endcsname
\providecommand{\newblock}{\relax}
\providecommand{\bibinfo}[2]{#2}
\providecommand{\BIBentrySTDinterwordspacing}{\spaceskip=0pt\relax}
\providecommand{\BIBentryALTinterwordstretchfactor}{4}
\providecommand{\BIBentryALTinterwordspacing}{\spaceskip=\fontdimen2\font plus
\BIBentryALTinterwordstretchfactor\fontdimen3\font minus
  \fontdimen4\font\relax}
\providecommand{\BIBforeignlanguage}[2]{{%
\expandafter\ifx\csname l@#1\endcsname\relax
\typeout{** WARNING: IEEEtranN.bst: No hyphenation pattern has been}%
\typeout{** loaded for the language `#1'. Using the pattern for}%
\typeout{** the default language instead.}%
\else
\language=\csname l@#1\endcsname
\fi
#2}}
\providecommand{\BIBdecl}{\relax}
\BIBdecl

\bibitem[Ni(2015)]{ni2015traffic}
D.~Ni, \emph{Traffic flow theory: Characteristics, experimental methods, and
  numerical techniques}.\hskip 1em plus 0.5em minus 0.4em\relax
  Butterworth-Heinemann, 2015.

\bibitem[An et~al.(2021)An, Guo, Hong, Lu, and Xia]{an2021lane}
C.~An, X.~Guo, R.~Hong, Z.~Lu, and J.~Xia, ``Lane-based traffic arrival pattern
  estimation using license plate recognition data,'' \emph{IEEE Intelligent
  Transportation Systems Magazine}, 2021.

\bibitem[Li et~al.(2023)Li, Tang, Chen, and Liu]{li2023traffic}
M.~Li, J.~Tang, Q.~Chen, and Y.~Liu, ``Traffic arrival pattern estimation at
  urban intersection using license plate recognition data,'' \emph{Physica A:
  Statistical Mechanics and its Applications}, vol. 625, p. 128995, 2023.

\bibitem[An et~al.(2024)An, He, Lu, Lu, and Xia]{an2024one}
C.~An, Y.~He, J.~Lu, Z.~Lu, and J.~Xia, ``One-stage estimation of cyclic
  arrival rates using license plate recognition data,'' \emph{Journal of
  Intelligent Transportation Systems}, pp. 1--14, 2024.

\bibitem[Zheng and Liu(2017)]{zheng2017estimating}
J.~Zheng and H.~X. Liu, ``Estimating traffic volumes for signalized
  intersections using connected vehicle data,'' \emph{Transportation Research
  Part C: Emerging Technologies}, vol.~79, pp. 347--362, 2017.

\bibitem[An et~al.(2018)An, Wu, Xia, and Huang]{an2018real}
C.~An, Y.-J. Wu, J.~Xia, and W.~Huang, ``Real-time queue length estimation
  using event-based advance detector data,'' \emph{Journal of Intelligent
  Transportation Systems}, vol.~22, no.~4, pp. 277--290, 2018.

\bibitem[Lee et~al.(2019)Lee, Xie, Ngoduy, and
  Keyvan-Ekbatani]{lee2019advanced}
S.~Lee, K.~Xie, D.~Ngoduy, and M.~Keyvan-Ekbatani, ``An advanced deep learning
  approach to real-time estimation of lane-based queue lengths at a signalized
  junction,'' \emph{Transportation research part C: emerging technologies},
  vol. 109, pp. 117--136, 2019.

\bibitem[Hao et~al.(2024)Hao, Lyuzhou, Oguchi, Keshuang, and
  Hong]{hao2024stochastic}
W.~Hao, L.~Lyuzhou, T.~Oguchi, T.~Keshuang, and Z.~Hong, ``Stochastic queue
  profile estimation using license plate recognition data,'' \emph{Physica A:
  Statistical Mechanics and its Applications}, vol. 643, p. 129790, 2024.

\bibitem[Li et~al.(2018)Li, Huang, and Lo]{li2018adaptive}
L.~Li, W.~Huang, and H.~K. Lo, ``Adaptive coordinated traffic control for
  stochastic demand,'' \emph{Transportation Research Part C: Emerging
  Technologies}, vol.~88, pp. 31--51, 2018.

\bibitem[Yao et~al.(2019)Yao, Shen, Liu, Jiang, and Yang]{yao2019dynamic}
Z.~Yao, L.~Shen, R.~Liu, Y.~Jiang, and X.~Yang, ``A dynamic predictive traffic
  signal control framework in a cross-sectional vehicle infrastructure
  integration environment,'' \emph{IEEE Transactions on Intelligent
  Transportation Systems}, vol.~21, no.~4, pp. 1455--1466, 2019.

\bibitem[Xu et~al.(2020)Xu, Wang, Wang, Li, Bertini, Qu, and
  Zhao]{xu2020trajectory}
Z.~Xu, Y.~Wang, G.~Wang, X.~Li, R.~L. Bertini, X.~Qu, and X.~Zhao, ``Trajectory
  optimization for a connected automated traffic stream: Comparison between an
  exact model and fast heuristics,'' \emph{IEEE Transactions on Intelligent
  Transportation Systems}, vol.~22, no.~5, pp. 2969--2978, 2020.

\bibitem[Yao et~al.(2020)Yao, Jiang, Cheng, Jiang, and Ran]{yao2020integrated}
Z.~Yao, H.~Jiang, Y.~Cheng, Y.~Jiang, and B.~Ran, ``Integrated schedule and
  trajectory optimization for connected automated vehicles in a conflict
  zone,'' \emph{IEEE Transactions on Intelligent Transportation Systems}, 2020.

\bibitem[Amini et~al.(2021)Amini, Omidvar, and
  Elefteriadou]{amini2021optimizing}
E.~Amini, A.~Omidvar, and L.~Elefteriadou, ``Optimizing operations at freeway
  weaves with connected and automated vehicles,'' \emph{Transportation Research
  Part C: Emerging Technologies}, vol. 126, p. 103072, 2021.

\bibitem[Dobrota et~al.(2022)Dobrota, Stevanovic, and
  Mitrovic]{dobrota2022novel}
N.~Dobrota, A.~Stevanovic, and N.~Mitrovic, ``A novel model to jointly estimate
  delay and arrival patterns by using high-resolution signal and detection
  data,'' \emph{Transportmetrica A: Transport Science}, pp. 1--33, 2022.

\bibitem[Tan et~al.(2021)Tan, Yao, Ban, and Tang]{tan2021cumulative}
C.~Tan, J.~Yao, X.~Ban, and K.~Tang, ``Cumulative flow diagram estimation and
  prediction based on sampled vehicle trajectories at signalized
  intersections,'' \emph{IEEE Transactions on Intelligent Transportation
  Systems}, 2021.

\bibitem[Luo et~al.(2019)Luo, Ma, Jin, Gong, and Wang]{luo2019queue}
X.~Luo, D.~Ma, S.~Jin, Y.~Gong, and D.~Wang, ``Queue length estimation for
  signalized intersections using license plate recognition data,'' \emph{IEEE
  Intelligent Transportation Systems Magazine}, vol.~11, no.~3, pp. 209--220,
  2019.

\bibitem[Tang et~al.(2022)Tang, Wu, Yao, Tan, and Ji]{tang2022lane}
K.~Tang, H.~Wu, J.~Yao, C.~Tan, and Y.~Ji, ``Lane-based queue length estimation
  at signalized intersections using single-section license plate recognition
  data,'' \emph{Transportmetrica B: transport dynamics}, vol.~10, no.~1, pp.
  293--311, 2022.

\bibitem[Shao and Chen(2018)]{shao2018license}
W.~Shao and L.~Chen, ``License plate recognition data-based traffic volume
  estimation using collaborative tensor decomposition,'' \emph{IEEE
  Transactions on Intelligent Transportation Systems}, vol.~19, no.~11, pp.
  3439--3448, 2018.

\bibitem[Rao et~al.(2018)Rao, Wu, Xia, Ou, and Kluger]{rao2018origin}
W.~Rao, Y.-J. Wu, J.~Xia, J.~Ou, and R.~Kluger, ``Origin-destination pattern
  estimation based on trajectory reconstruction using automatic license plate
  recognition data,'' \emph{Transportation Research Part C: Emerging
  Technologies}, vol.~95, pp. 29--46, 2018.

\bibitem[Mo et~al.(2020)Mo, Li, and Dai]{mo2020estimating}
B.~Mo, R.~Li, and J.~Dai, ``Estimating dynamic origin--destination demand: A
  hybrid framework using license plate recognition data,'' \emph{Computer-Aided
  Civil and Infrastructure Engineering}, vol.~35, no.~7, pp. 734--752, 2020.

\bibitem[Yao et~al.(2021)Yao, Zhang, Jin, and Ma]{yao2021understanding}
W.~Yao, M.~Zhang, S.~Jin, and D.~Ma, ``Understanding vehicles commuting pattern
  based on license plate recognition data,'' \emph{Transportation Research Part
  C: Emerging Technologies}, vol. 128, p. 103142, 2021.

\bibitem[Zhan et~al.(2015)Zhan, Li, and Ukkusuri]{zhan2015lane}
X.~Zhan, R.~Li, and S.~V. Ukkusuri, ``Lane-based real-time queue length
  estimation using license plate recognition data,'' \emph{Transportation
  Research Part C: Emerging Technologies}, vol.~57, pp. 85--102, 2015.

\bibitem[Mo et~al.(2017)Mo, Li, and Zhan]{mo2017speed}
B.~Mo, R.~Li, and X.~Zhan, ``Speed profile estimation using license plate
  recognition data,'' \emph{Transportation research part C: emerging
  technologies}, vol.~82, pp. 358--378, 2017.

\bibitem[Day et~al.(2010)Day, Haseman, Premachandra, Brennan~Jr, Wasson,
  Sturdevant, and Bullock]{day2010evaluation}
C.~M. Day, R.~Haseman, H.~Premachandra, T.~M. Brennan~Jr, J.~S. Wasson, J.~R.
  Sturdevant, and D.~M. Bullock, ``Evaluation of arterial signal coordination:
  Methodologies for visualizing high-resolution event data and measuring travel
  time,'' \emph{Transportation Research Record}, vol. 2192, no.~1, pp. 37--49,
  2010.

\bibitem[Zheng et~al.(2014)Zheng, Liu, Misgen, Schwartz, Green, and
  Anderson]{zheng2014use}
J.~Zheng, H.~X. Liu, S.~Misgen, K.~Schwartz, B.~Green, and M.~Anderson, ``Use
  of event-based traffic data in generating time--space diagrams for evaluation
  of signal coordination,'' \emph{Transportation Research Record}, vol. 2439,
  no.~1, pp. 94--104, 2014.

\bibitem[Zhan et~al.(2016)Zhan, Zheng, Yi, and Ukkusuri]{zhan2016citywide}
X.~Zhan, Y.~Zheng, X.~Yi, and S.~V. Ukkusuri, ``Citywide traffic volume
  estimation using trajectory data,'' \emph{IEEE Transactions on Knowledge and
  Data Engineering}, vol.~29, no.~2, pp. 272--285, 2016.

\bibitem[Zhao et~al.(2019)Zhao, Zheng, Wong, Wang, Meng, and
  Liu]{zhao2019various}
Y.~Zhao, J.~Zheng, W.~Wong, X.~Wang, Y.~Meng, and H.~X. Liu, ``Various methods
  for queue length and traffic volume estimation using probe vehicle
  trajectories,'' \emph{Transportation Research Part C: Emerging Technologies},
  vol. 107, pp. 70--91, 2019.

\bibitem[Hao et~al.(2013)Hao, Sun, Ban, Guo, and Ji]{hao2013vehicle}
P.~Hao, Z.~Sun, X.~J. Ban, D.~Guo, and Q.~Ji, ``Vehicle index estimation for
  signalized intersections using sample travel times,'' \emph{Transportation
  Research Part C: Emerging Technologies}, vol.~36, 2013.

\bibitem[Zhang et~al.(2019)Zhang, Liu, Chen, Yu, and Wang]{zhang2019cycle}
H.~Zhang, H.~X. Liu, P.~Chen, G.~Yu, and Y.~Wang, ``Cycle-based end of queue
  estimation at signalized intersections using low-penetration-rate vehicle
  trajectories,'' \emph{IEEE Transactions on Intelligent Transportation
  Systems}, vol.~21, no.~8, pp. 3257--3272, 2019.

\bibitem[Li et~al.(2017)Li, Tang, Yao, and Li]{li2017real}
F.~Li, K.~Tang, J.~Yao, and K.~Li, ``Real-time queue length estimation for
  signalized intersections using vehicle trajectory data,''
  \emph{Transportation Research Record}, vol. 2623, no.~1, pp. 49--59, 2017.

\bibitem[Wen and Weng(2023)]{wen2023inferring}
Z.~Wen and X.~Weng, ``Inferring the number of vehicles between
  trajectory-observed vehicles,'' \emph{Journal of Intelligent Transportation
  Systems}, pp. 1--14, 2023.

\bibitem[Blundell et~al.(2015)Blundell, Cornebise, Kavukcuoglu, and
  Wierstra]{blundell2015weight}
C.~Blundell, J.~Cornebise, K.~Kavukcuoglu, and D.~Wierstra, ``Weight
  uncertainty in neural network,'' in \emph{International conference on machine
  learning}.\hskip 1em plus 0.5em minus 0.4em\relax PMLR, 2015, pp. 1613--1622.

\end{thebibliography}

\vspace{-12 mm} 
\begin{IEEEbiography}[{\includegraphics[width=1in,height=1.25in,clip,keepaspectratio]{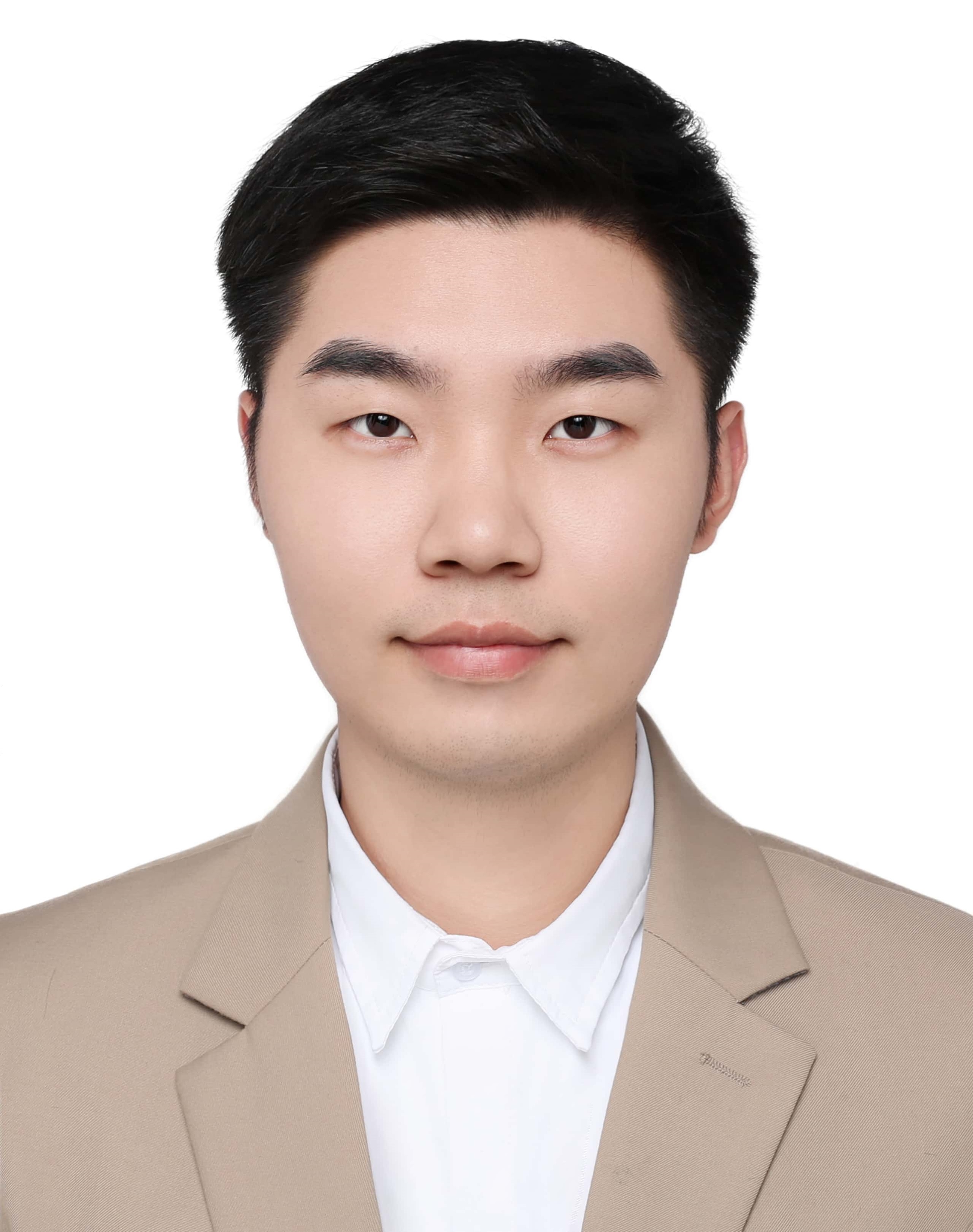}}]{Yang He}
    received his B.E. degree in Traffic Engineering from Chang’ an University, Xi'an, China, in 2020. He is currently pursuing the Ph.D. degree in the Intelligent Transportation System Research Center at Southeast University. His current research interests include transportation network modeling, traffic state estimation, and low-rank modeling.
    
\end{IEEEbiography}

\begin{IEEEbiography}[{\includegraphics[width=1in,height=1.25in,clip,keepaspectratio]{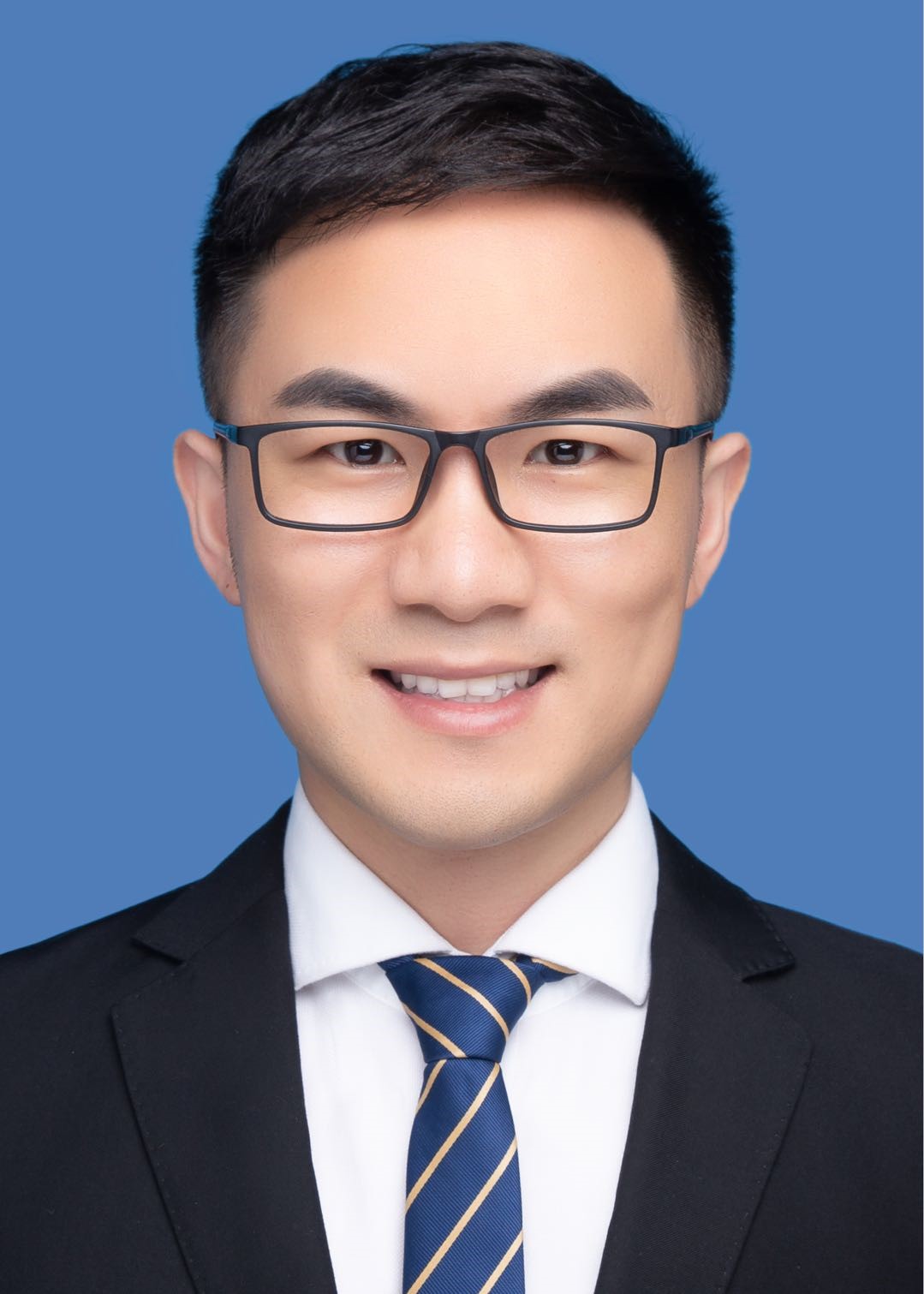}}]{Chengchuan An}
    received his Ph.D. degree in Transportation Engineering from Southeast University, Nanjing, China, in 2019. From 2014 to 2016, he was a visiting scholar at the Department of Civil and Architectural Engineering and Mechanics, University of Arizona, USA. Since 2020, he has been a post-doctor in the Intelligent Transportation System Research Center at Southeast University. His current research interests include intelligent traffic signal control systems and traffic data mining.
\end{IEEEbiography}

\begin{IEEEbiography}[{\includegraphics[width=1in,height=1.25in,clip,keepaspectratio]{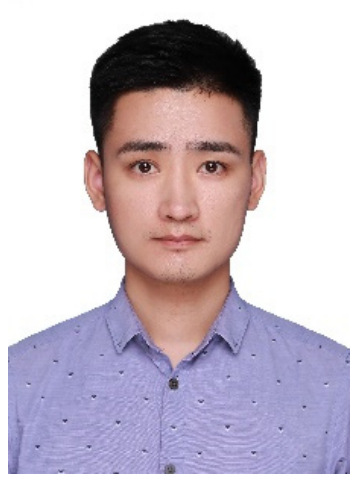}}]{Jiawei Lu}
  received his Ph.D. degree in transportation engineering from Arizona State University in 2022. He is currently a Postdoctoral Fellow with the H. Milton Stewart School of Industrial and Systems Engineering at Georgia Institute of Technology. His research interests include transportation network modeling, traffic simulation, and distributed traffic control.
\end{IEEEbiography}


\begin{IEEEbiography}[{\includegraphics[width=1in,height=1.2in,clip,keepaspectratio]{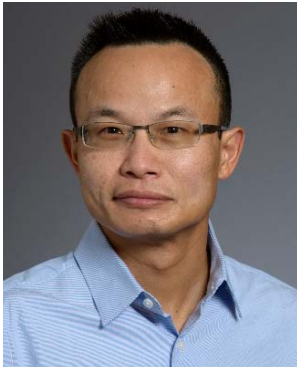}}]{Yao-Jan Wu} is a Professor in transportation engineering with the Civil and Architectural Engineering and Mechanics Department and the Executive (Founding) Director of the Center for Applied Transportation Sciences (CATS), The University of Arizona (UA). Dr. Wu has authored or co-authored over 160 refereed publications, including more than 80 journal articles, and has presented his research findings at more than 100 national and international conferences and invited speaker events. His research focuses on a strong connection between information technology (IT) and traditional transportation research.

\end{IEEEbiography}

\begin{IEEEbiography}[{\includegraphics[width=1in,height=1.25in,clip,keepaspectratio]{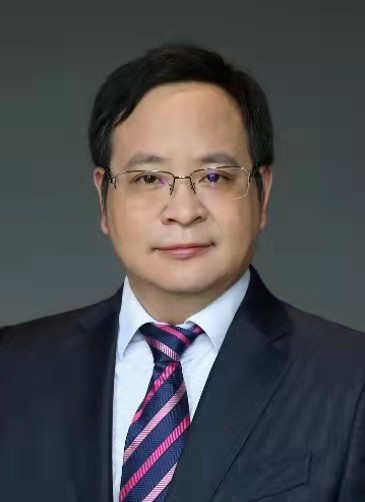}}]{Zhenbo Lu}
    received the Ph.D. degree in Traffic Information Engineering and Control from Southeast University, Nanjing, China, in 2011. He is currently an Associate Professor with the Intelligent Transportation System Research Center, Southeast University. His main research interests include transportation planning, traffic simulation, and intelligent transportation systems.
\end{IEEEbiography}
\begin{IEEEbiography}[{\includegraphics[width=1in,height=1.25in,clip,keepaspectratio]{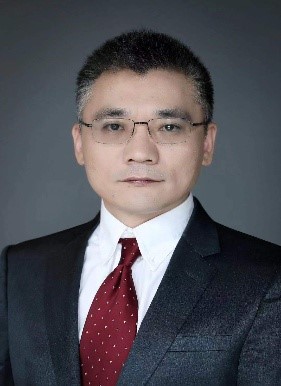}}]{Jingxin Xia}
is a professor at the Intelligent Transportation System Research Center, Southeast University, Nanjing, China. He received the Ph.D. degree in Transportation Engineering from the University of Kentucky, USA in 2006. He has published more than forty peer-reviewed papers so far, and his main research interests include traffic flow theory, transportation network modeling, traffic signal control, and intelligent transportation systems.
\end{IEEEbiography}

\end{document}